\documentclass{article}

\PassOptionsToPackage{square,numbers, sort&compress}{natbib}

\usepackage[main,final]{neurips_2026}

\usepackage[utf8]{inputenc} %
\usepackage[T1]{fontenc}    %
\usepackage{hyperref}       %
\hypersetup{hidelinks}
\usepackage{url}            %
\usepackage{booktabs}       %
\usepackage{amsfonts}       %
\usepackage{nicefrac}       %
\usepackage{microtype}      %
\usepackage[table]{xcolor}  %
\usepackage{cancel}
\usepackage{graphicx}
\usepackage{subcaption}
\usepackage{float}
\usepackage{adjustbox}
\usepackage{multirow}
\usepackage{makecell}
\usepackage{tikz}
\usetikzlibrary{arrows.meta,calc,fit,positioning,shapes.geometric,shapes.misc,backgrounds}
\usepackage{amssymb}
\usepackage{wrapfig,lipsum,booktabs}
\usepackage{xr}
\usepackage{xspace}
\usepackage{amssymb}%
\usepackage{pifont}%
\usepackage{microtype}

\newcommand{\newparagraph}[1]{\noindent \textbf{#1}}

\definecolor{workingblue}{HTML}{4472C4}
\definecolor{anchorred}{HTML}{C0392B}
\definecolor{pairAbase}{HTML}{C4DAFC}   %
\definecolor{pairAours}{HTML}{E5EFFD}   %
\definecolor{pairBbase}{HTML}{F5CDCD}   %
\definecolor{pairBours}{HTML}{FCE9E9}   %
\definecolor{sflight}{HTML}{EBF3FD}    %
\definecolor{sfmid}{HTML}{C4DAFC}      %
\definecolor{sfdark}{HTML}{9BB8F7}     %
\definecolor{cflight}{HTML}{FCE9E9}    %
\definecolor{cfmid}{HTML}{F5CDCD}      %
\definecolor{cfdark}{HTML}{E88888}     %
\definecolor{groupgray}{HTML}{EFEFEF} %
\newcommand{\workingheads}{\textcolor{workingblue}{\textbf{Working Heads}}}
\newcommand{\anchorheads}{\textcolor{anchorred}{\textbf{Anchor Heads}}}
\definecolor{mediumslateblue}{rgb}{0.43, 0.43, 1}
\definecolor{salmon}{rgb}{1, 0.42, 0.42}
\definecolor{iris}{rgb}{0.35, 0.31, 0.81}

\usepackage{csquotes}

\usepackage{xspace}
\usepackage{amsmath}  %

\makeatletter
\DeclareRobustCommand\onedot{\futurelet\@let@token\@onedot}
\def\@onedot{\ifx\@let@token.\else.\null\fi\xspace}

\def\eg{e.g\onedot} 
\def\ie{i.e\onedot}

\makeatother

\newcommand{\bx}{{\mathbf{x}}}

\newcommand{\bz}{{\mathbf{z}}}
\newcommand{\bA}{{\mathbf{A}}}

\newcommand{\bK}{{\mathbf{K}}}

\newcommand{\bQ}{{\mathbf{Q}}}

\newcommand{\bV}{{\mathbf{V}}}

\newcommand{\cA}{\mathcal{A}}

\usepackage[capitalize]{cleveref}

\title{Disen-Forcing: Disentangling Semantic Anchoring from Motion for Autoregressive Video Diffusion}

\author{%
\textbf{Zhengyang Yu}$^{1}$ \quad \textbf{Akio Hayakawa}$^{2}$ \quad
\textbf{Masato Ishii}$^{2}$ \quad \textbf{Qingtao Yu}$^{1}$ \\
\textbf{Xuesong Li}$^{1,3}$ \quad \textbf{Takashi Shibuya}$^{2}$ \quad
\textbf{Yuki Mitsufuji}$^{2,4}$ \quad \textbf{Jing Zhang}$^{1}$ \\
$^{1}$Australian National University \quad $^{2}$Sony AI \\
$^{3}$CSIRO \quad $^{4}$Sony Group Corporation \\
{\small\texttt{\{zhengyang.yu,terry.yu,xuesong.li,jing.zhang\}@anu.edu.au}} \\
{\small\texttt{\{akio.hayakawa,masato.a.ishii,takashi.tak.shibuya,yuhki.mitsufuji\}@sony.com}} \\[4pt]
{\small \textbf{Project page:} \href{https://disen-forcing.github.io/}{\texttt{\textcolor[HTML]{C0392B}{h}\textcolor[HTML]{BC3B34}{t}\textcolor[HTML]{B93E3D}{t}\textcolor[HTML]{B54045}{p}\textcolor[HTML]{B2434E}{s}\textcolor[HTML]{AE4557}{:}\textcolor[HTML]{AA4860}{/}\textcolor[HTML]{A74A69}{/}\textcolor[HTML]{A34C72}{d}\textcolor[HTML]{A04F7A}{i}\textcolor[HTML]{9C5183}{s}\textcolor[HTML]{99548C}{e}\textcolor[HTML]{955695}{n}\textcolor[HTML]{91599E}{-}\textcolor[HTML]{8E5BA6}{f}\textcolor[HTML]{895CA8}{o}\textcolor[HTML]{855EAA}{r}\textcolor[HTML]{815FAB}{c}\textcolor[HTML]{7C60AD}{i}\textcolor[HTML]{7862AF}{n}\textcolor[HTML]{7463B1}{g}\textcolor[HTML]{6F65B2}{.}\textcolor[HTML]{6B66B4}{g}\textcolor[HTML]{6767B6}{i}\textcolor[HTML]{6269B8}{t}\textcolor[HTML]{5E6AB9}{h}\textcolor[HTML]{5A6BBB}{u}\textcolor[HTML]{556DBD}{b}\textcolor[HTML]{516EBF}{.}\textcolor[HTML]{4D6FC0}{i}\textcolor[HTML]{4871C2}{o}\textcolor[HTML]{4472C4}{/}}}}
}

\begin{document}
\setlength{\intextsep}{18pt plus 2pt minus 2pt}

\maketitle
\vspace{-8pt} 

\begin{center}
\begin{minipage}{\linewidth}
\centering
\includegraphics[width=\linewidth]{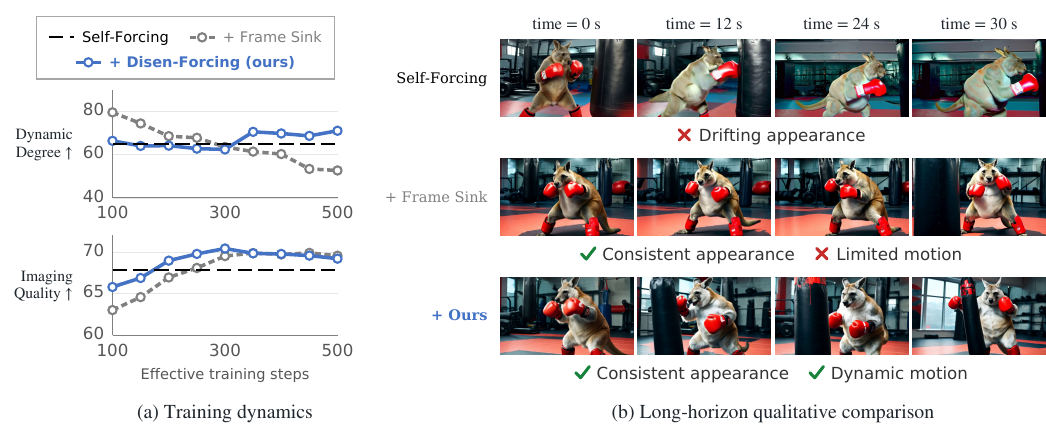}
\captionof{figure}{\textbf{Training dynamics and long-horizon comparison.} (a) Dynamic Degree over the first 5s and Imaging Quality over 30s across 5s-horizon training iterations. (b) Qualitative comparison of 30s generations. Self-Forcing exhibits appearance drift, while Frame Sink preserves appearance but limits motion dynamics. Disen-Forcing maintains both consistent appearance and dynamic motion.}
\label{fig:teaser}
\end{minipage}
\end{center}

\begin{abstract}
Autoregressive video diffusion models (AR-VDMs) enable real-time, long-horizon video generation but degrade rapidly once the inference horizon outruns the training horizon, due to compounding exposure bias. To stabilize long rollouts, recent methods heuristically adopt \emph{frame sink}---keeping initial frames as a persistent model context, inspired by attention sinks in LLMs. We revisit this design and find that the mechanism by which initial frames stabilize long-range consistency in AR-VDMs is \emph{not} an LLM-style attention sink; instead, long-range consistency is anchored through a sparse subset of attention heads, while the majority of heads predominantly capture local temporal dynamics. This observation reveals an over-conditioning pathology in existing frame-sink strategies: by universally exposing every head to the pinned initial-frame context, the model becomes prone to collapsing motion toward the low-dynamic modes of the teacher during model distillation. We propose \textbf{Disen-Forcing}, a distillation-time plug-in for AR-VDMs that disentangles and reinforces this head specialization through \textit{disentangled context routing} and \textit{motion-disentangling regularization}. On long-horizon video generation benchmarks, Disen-Forcing simultaneously improves long-range consistency and motion dynamics of base AR-VDMs---a trade-off that sink-based methods fail to balance.

\end{abstract}
\section{Introduction}
\label{sec:intro}
Recent advances in video diffusion models~(VDMs) have enabled photorealistic and temporally coherent video synthesis~\cite{ho2022video,hunyuan,chen2023videocrafter1,chen2024videocrafter2}. However, most VDMs employ bidirectional temporal attention, generating all frames in parallel without respecting temporal order, which limits their use in real-time or interactive applications. Autoregressive video diffusion models (AR-VDMs)~\cite{diffusionforcing,causvid,selfforcing} have emerged as a promising alternative, applying per-frame diffusion denoising in causal order. This hybrid paradigm retains diffusion's high-fidelity sample quality while enabling streaming generation, a key capability for real-time interactive world models~\cite{genie3,yume,lingbot,Hunyuan-GameCraft}.

A fundamental challenge in AR-VDMs is \emph{exposure bias}~\cite{schmidt2019generalization,framepack}~(\cref{sec:preliminary}): as each newly generated frame is fed back as future conditioning, small generation errors accumulate over rollout steps, progressively degrading visual quality and semantic consistency. This issue is exacerbated by a \emph{training--inference mismatch} in horizon and context quality: training typically exposes the model to limited-length rollouts and relatively clean context, whereas inference often requires much longer extrapolation under increasingly error-corrupted self-generated history. Self-Forcing~\cite{selfforcing} mitigates exposure bias by training on self-generated rollouts, but drift still emerges when the inference horizon exceeds the training horizon. To solve this, the de facto stabilization heuristic is \emph{frame sink}~\cite{longlive,rollingforcing,deepforcing} (\cref{sec:preliminary}), which pins the initial frames in the KV cache throughout model rollout, inspired by attention sinks in streaming LLMs~\cite{streamingllm}---a design that implicitly treats initial frames as a universal attention attractor, i.e., a shared global reference for all heads. While it improves long-range semantic consistency, it often suppresses temporal dynamics, yielding conservative motion (\cref{sec:collapse}).

In this work, by analyzing the temporal attention patterns of AR-VDMs (\cref{sec:uncover}), we find that initial frames do not act as a universal attention attractor as in LLM-style attention sinks; instead, they anchor semantic consistency through a sparse, stable subset of \emph{anchor heads}. This reveals an over-conditioning in the existing frame sink strategy, which universally exposes every head to the initial-frame context. Under the mode-seeking distillation objective~\cite{dmd}, this universalization causes the student to collapse motion toward the low-variance, near-static modes of the teacher (\cref{sec:collapse}). 

Building on these observations, we propose \textbf{Disen-Forcing}, a distillation-time plug-in for AR-VDMs that disentangles and reinforces the head specialization through two complementary components: \textit{disentangled context routing} (\cref{sec:contextsep}) replaces the universal sink during DMD with per-head-type contexts: only anchor heads retain access to the initial frames, while the rest are restricted to the local sliding window, eliminating the over-conditioning; \textit{motion-disentangling regularization} (\cref{sec:reg}) imposes an invariance constraint preventing anchor heads from leaking global semantics into motion, directly suppressing the remaining semantic-to-motion leakage after routing. Our contributions are:
\begin{itemize}
    \item \textbf{Diagnosis:} revealing that initial frames in AR-VDMs do not act as universal attention sinks; instead, semantic anchoring is concentrated in a sparse, stable subset of \emph{anchor heads}, exposing universal frame-sink conditioning as a cause of motion collapse under DMD.
    \item \textbf{Method:} proposing \textbf{Disen-Forcing}, a distillation-time plug-in that reinforces this head specialization via \emph{disentangled context routing} and \emph{motion-disentangling regularization}.
    \item \textbf{Results:} demonstrating on long-horizon video benchmarks that Disen-Forcing improves both motion dynamics and long-range consistency over base AR-VDMs, resolving a trade-off that existing sink-based methods cannot balance.
\end{itemize}

\section{Related Work}
\label{sec:related_work}
\newparagraph{AR-VDMs.} 
Video diffusion models (VDMs)~\cite{he2022latent,guo2023animatediff,wang2023modelscopetexttovideotechnicalreport,wang2023lavie,chen2023videocrafter1,chen2024videocrafter2,Open-Sora,mochi,hunyuan,yang2024cogvideox,wan2025wan} generate high-fidelity content with coherent temporal dynamics. Autoregression (AR), established in language modeling~\cite{brown2020language} and token-based video generation~\cite{hong2022cogvideo}, also enables incremental video synthesis. Rolling-denoising methods~\cite{rollingdiffusion,rollingforcing,fifo,kodaira2025streamdit,pavdm,Ardiffusion} progressively denoise and advance a temporal window, whereas FramePack~\cite{framepack} compresses historical context for next-frame-section prediction. Diffusion Forcing~\cite{diffusionforcing} trains with independent per-frame noise levels; History Guidance~\cite{song2025history} extends this formulation to non-causal transformers. SkyReels-V2~\cite{skyreels} and MAGI-1~\cite{magi} use ordered frame- or chunk-wise noise schedules for long-video generation.
Within this diffusion-AR hybrid paradigm, CausVid~\cite{causvid} attains few-step causal inference through Distribution Matching Distillation (DMD)~\cite{dmd,dmd2}, while Self-Forcing~\cite{selfforcing} retains the DMD objective and narrows the train--test gap by feeding the model's own rollout as context during training. Causal-Forcing~\cite{causalforcing} further improves initialization by distilling ODE trajectories from an autoregressive teacher. Despite training on self-generated context, temporal drift still emerges when inference exceeds the training horizon.

\newparagraph{Attention Sinks in LLMs.}
The attention sink phenomenon~\cite{streamingllm} in AR LLMs---disproportionate attention to the first few input tokens regardless of semantic content---emerges with large-scale next-token prediction training~\cite{attentionsink2}. These tokens act as no-op attention score absorbers~\cite{attentionsink}; pinning them as persistent context stabilizes streaming inference over arbitrarily long sequences~\cite{streamingllm}.

\newparagraph{Sink-based AR-VDMs.}
Inspired by LLM attention sinks, the frame sink heuristic~\cite{longlive,rollingforcing,deepforcing,rollingsink,FLEX,lol,infiniteforcing} has been widely adopted in AR-VDMs to alleviate long-horizon drift by pinning initial frames as persistent context. We show that, in AR-VDMs, sink-like attention to initial frames is not ubiquitous across heads but concentrated in a sparse subset of heads (\cref{sec:uncover}); universally exposing every head to the pinned context drives motion collapse under mode-seeking distillation (\cref{sec:collapse}).

\section{Rethinking the Frame Sink Strategy in AR-VDMs}
In this section, we begin by revisiting AR-VDMs, their exposure bias, and the frame sink heuristic (\cref{sec:preliminary}). Then we show that initial frames are not a universal attention sink but a clean reference exploited by a sparse subset of attention heads for global consistency (\cref{sec:uncover}). Finally, we identify motion collapse as the consequence of universalizing this role across all heads (\cref{sec:collapse}).
\subsection{Preliminaries: AR-VDMs and Anti-Drift with Frame Sink}
\label{sec:preliminary}
\newparagraph{Exposure Bias in AR-VDMs.} An AR-VDM is a hybrid model that combines the strengths of AR models in temporal modeling and diffusion models in continuous distribution modeling. Specifically, it uses the AR chain-rule decomposition to factorize the joint distribution of a video sequence $\bx^{1:N} = (\bx^1, \bx^2, \ldots, \bx^N)$ into a product of conditionals\footnote{Each $\bx^i$ may represent a single frame or a chunk of frames; we use ``frame'' and ``chunk'' interchangeably for brevity.}:
$p(\bx^{1:N}) = \prod_{i=1}^N p(\bx^i \mid \bx^{<i})$. Each conditional $p(\bx^i \mid \bx^{<i})$ is modeled by a diffusion process that denoises a Gaussian noise into a video frame conditioned on previously generated frames. These conditionals can be learned with denoising~\cite{diffusionforcing} or flow-matching~\cite{lipman2022flow} objectives. In Teacher Forcing, the conditional at noise level $t_j$ is formed as $p(\bx_{t_j}^i \mid \bx_0^{<i})$, with clean ground-truth history frames. In Diffusion Forcing~\cite{diffusionforcing}, the conditional is instead $p(\bx_{t_j}^i \mid \bx_{t_{\geq 0}}^{<i})$, where history frames are corrupted with independent noise levels. Both formulations introduce a train--test gap termed \emph{exposure bias}~\cite{schmidt2019generalization}: the context exposed to the model is ground-truth during training but self-generated at inference.

\begin{figure}[!htbp]
\centering
\resizebox{\textwidth}{!}{%
\includegraphics[width=\linewidth]{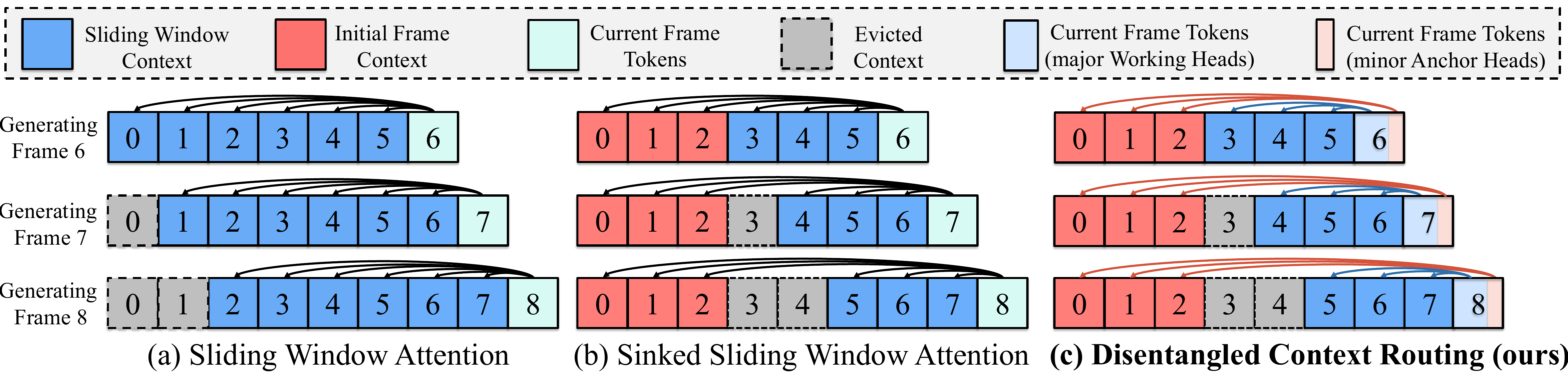}%
}
\caption{\textbf{Disentangled context routing.} (a)~Sliding window attention loses early frames as the cache rolls forward. (b)~Sinked sliding window retains initial frames but routes the same context to all heads. (c)~During DMD, our approach routes initial-frame context exclusively to the sparse \anchorheads{} while the majority \workingheads{} attend only to the local sliding window, preserving the natural head specialization.}
\vspace{-1.0em}
\label{fig:method}
\end{figure}

\newparagraph{Causal Distillation.} As an alternative to the above methods, CausVid~\cite{causvid} shifts toward causal distillation. This is typically achieved through Distribution Matching Distillation~\cite{dmd} (DMD). Rather than training with a per-frame denoising objective on ground-truth videos, a causal student model $G_\theta$, initialized by regression on the bidirectional teacher's ODE trajectories~\cite{causvid,selfforcing}, is distilled to directly build a noise$\to$data mapping, using the holistic fidelity of its generated videos under the teacher's distribution as its learning signal. Concretely, DMD minimizes the reverse KL divergence between the student distribution $q_{\theta,t}$ and the teacher distribution $p_t$ at diffusion timestep $t$~\cite{vsd,dmd}, with gradient
\begin{equation}
\nabla_\theta \mathcal{L}_{\mathrm{DMD}} = \mathbb{E}_t\bigl[\nabla_\theta \operatorname{KL}(q_{\theta,t} \,\|\, p_t)\bigr].
\label{eq:dmd}
\end{equation}
Training with the distribution-matching loss enables few-step sampling without paired teacher trajectories. Self-Forcing~\cite{selfforcing} further mitigates the train--test gap by conditioning the model on self-rollout frames during distillation. Self-Forcing implements history conditioning through causal self-attention with a KV cache~\cite{causvid,selfforcing}. Since the context $\bx^{<i}$ grows unboundedly with AR steps, Self-Forcing adopts sliding window attention (\cref{fig:method}~(a)), which is also referred to as rolling KV cache, approximating $p(\bx^i \mid \bx^{<i})$ by $q_\theta(\bx^i \mid \bx^{i-w:i-1}) $, retaining only the $w$ most recent frames. Although self-rollout training alleviates exposure bias, severe error compounding still occurs once inference extrapolates beyond the training horizon.

\newparagraph{Anti-Drift with Frame Sink.}
Recent works~\cite{rollingforcing,longlive} empirically identify that retaining initial frames as a ``frame sink''~\cite{longlive} in the sliding window attention, \ie, $q_\theta(\bx^i \mid \bx^1, \bx^{i-w:i-1})$---a strategy analogous to the attention sinks in LLMs~\cite{streamingllm}, is effective at maintaining long-range semantic consistency during frame-length extrapolation. While this heuristic mitigates appearance drift from error accumulation, the role of initial frames in preventing drift remains unclear. In LLMs, the attention sink refers to a phenomenon where most heads assign disproportionate attention mass to the first few tokens, which act as ``no-ops'' absorbing probability mass in the absence of relevant context (\cref{sec:related_work}). A natural question arises: \emph{do initial frame tokens serve the same role in AR-VDMs?}

\begin{figure}[H]
    \centering
    \begin{subfigure}{\linewidth}
        \centering
        \includegraphics[width=0.95\linewidth]{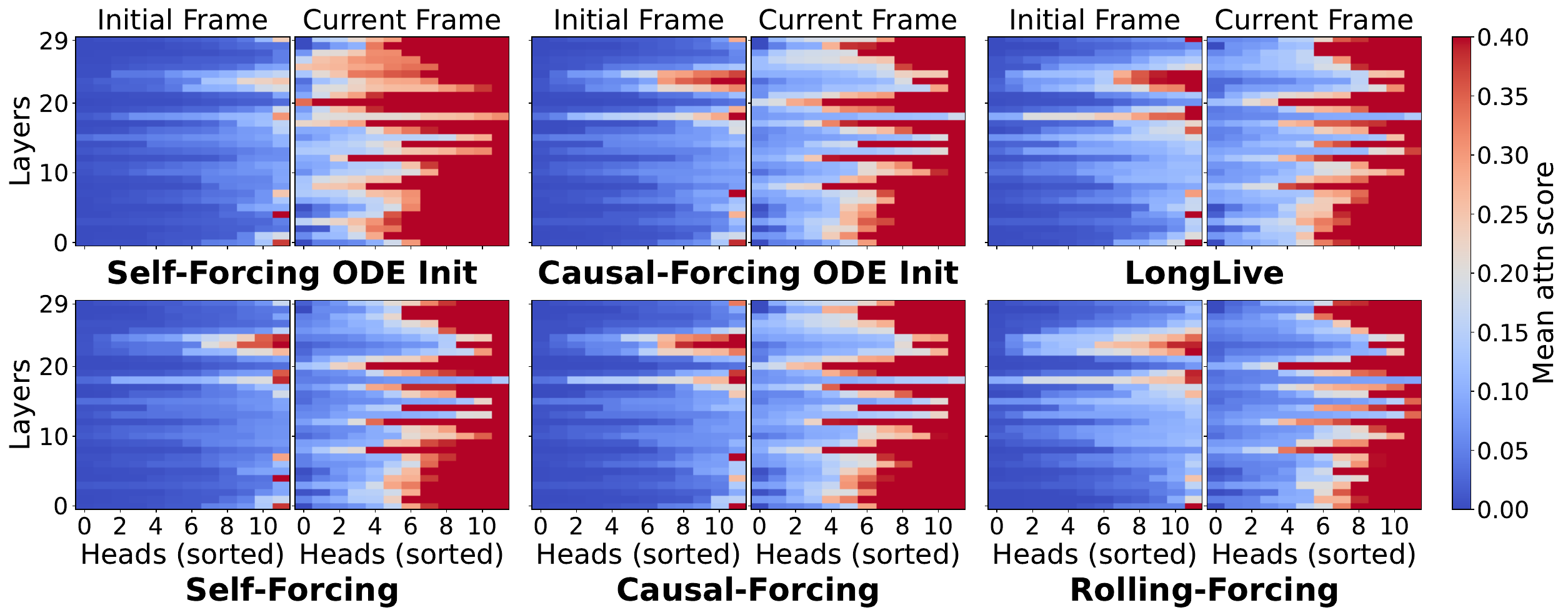}
        \caption{Per-head attention across AR-VDMs}\label{fig:sink_value}
    \end{subfigure}
    \par\smallskip
    \begin{subfigure}{\linewidth}
        \centering
        \includegraphics[width=0.95\linewidth]{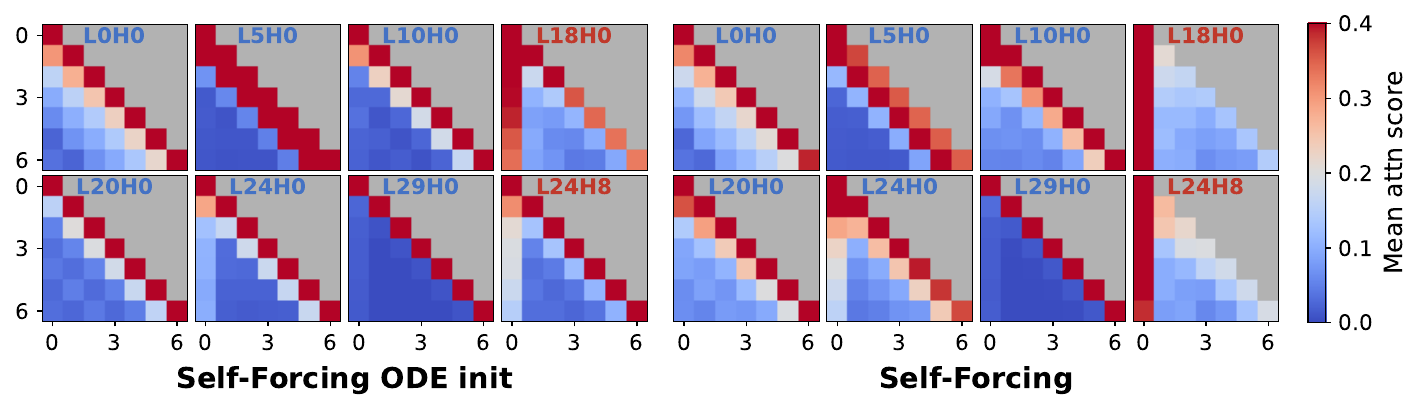}
        \caption{Representative Self-Forcing heads over rollout steps}\label{fig:headwise}
    \end{subfigure}
    \caption{\textbf{Temporal attention patterns across models and heads.}
    (a) Per-head attention to the initial and current frames across AR-VDMs, averaged over 50 MovieGenBench~\cite{moviegen} prompts and denoising timesteps. Most heads attend predominantly to the current frame, while a sparse subset concentrates attention on the initial frame.
    (b) Attention distributions of representative Self-Forcing heads across seven rollout steps, at ODE initialization (left) and after DMD (right). \workingheads{} exhibit local temporal patterns, whereas \anchorheads{} attend persistently to the initial frame.}
    \label{fig:temporal_attention}
    \vspace{-1em}
\end{figure}

\subsection{Uncovering the Anti-Drift Role of Initial Frames}
\label{sec:uncover}
\paragraph{Initial Frames Are Not a Universal Attention Sink.}
To uncover the mechanism of initial frames in anti-drift, we examine per-head attention scores across layers and AR steps. We observe that the attention landscape in AR-VDMs is fundamentally different from that in LLMs. Specifically, we measure the average attention allocated to the initial frames versus the current frame for every head, averaged over 50 MovieGenBench~\cite{moviegen} prompts and all diffusion timesteps. As shown in Fig.~\ref{fig:sink_value}, the vast majority of heads attend predominantly to the current frame---not the initial frames. In other words, \emph{the current frame, rather than the initial ones, plays a role similar to the universal attention sink in LLMs}. We conjecture that the difference between the LLM and AR-VDM sink patterns stems from two factors. \emph{Data:} in language, adjacent tokens are discrete and typically less redundant than adjacent video frames, which often share substantial visual content. \emph{Objective:} In LLMs, the prediction target is a future token outside the input; under causal masking, initial tokens are uniquely visible to every query and naturally absorb the resulting attention mass as sinks~\cite{streamingllm}. AR-VDMs, in contrast, denoise the current frame from its own noisy latent at each AR step: the denoising target is located at the current-frame position itself. Together with the strong spatio-temporal redundancy of video tokens, this encourages attention to concentrate on the current frame.

\begin{figure}[!htbp]
    \centering
    \includegraphics[width=0.85\linewidth]{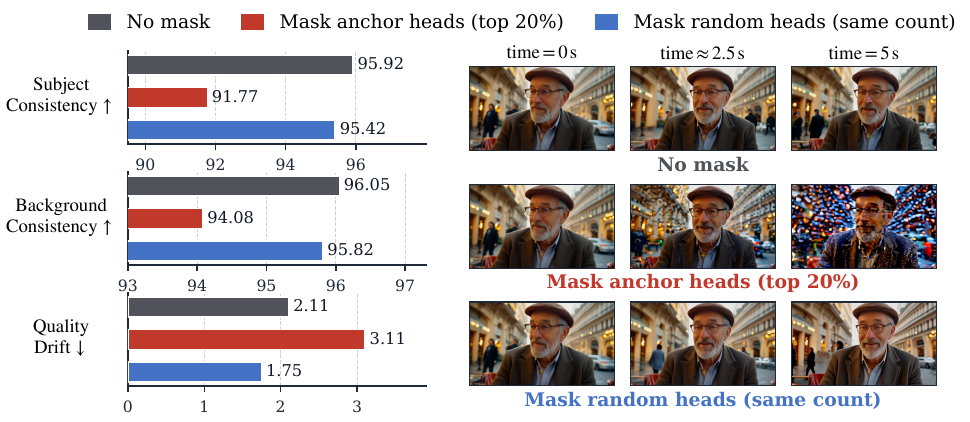}
\caption{\textbf{Critical Role of Anchor Heads in Semantic Consistency.}
Masking the top 20\% \anchorheads\ leads to a sharp drop in subject/background consistency and a clear rise in imaging-quality drift (left), together with visible appearance drift \textbf{even within the 5-second training horizon} (right). Masking the same number of \workingheads\ at random yields metrics on par with the unmasked baseline and no observable drift.}
\vspace{-1.5em}
    \label{fig:mask_head}
\end{figure}

\newparagraph{Initial Frames Sustain Consistency through Sparse Heads.}
To discover what truly enables initial frames in anti-drift, we visualize per-head attention distributions throughout the AR rollout (Fig.~\ref{fig:headwise}; representative head-wise attention grids in \cref{fig:head_full_grid}). While most heads exhibit a local temporal pattern, a sparse subset, concentrated in specific layers (\eg, layers 18 and 22--24 of the 30-layer Wan2.1~\cite{wan2025wan} backbone), exhibits persistent attention spikes toward the initial frame. Based on this observation, we categorize attention heads into two types:
\workingheads---The vast majority of heads, which attend predominantly to the local sliding window context; and \anchorheads---A sparse minority of heads that attend strongly to the initial frame across AR steps.

A universal attention sink would absorb the majority of the attention mass across heads; the observed sparsity of \anchorheads{} suggests a different mechanism. From the attention pattern, we hypothesize that these sparse \anchorheads{} exploit the initial frames as a clean reference with minimal cumulative error, stabilizing long-range consistency. We test this via a head-masking experiment (\cref{fig:mask_head}): during inference within the 5-second Self-Forcing training horizon, we mask the top-20\% anchor heads (by average initial-frame attention score). This causes visible appearance drift \emph{even within the training horizon}. In contrast, masking the same number of randomly selected \workingheads{} produces no observable drift (more samples in \cref{fig:head_masking_qual}). Together, the two head types reveal a division of labor: \anchorheads{} sustain global semantic consistency by preserving reference to the initial frames, while \workingheads{} focus on local temporal dynamics, as reflected in their local attention pattern.

\begin{figure}[!htbp]
    \centering
    \includegraphics[width=0.90\linewidth]{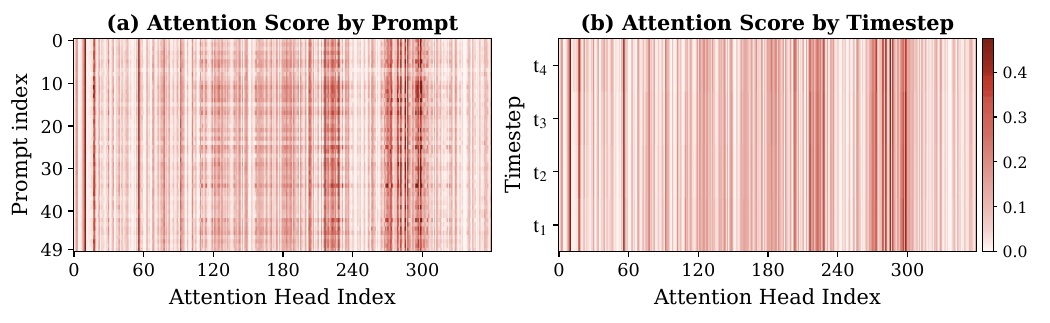}
\caption{\textbf{Anchor-head positions are input invariant.} (a) Each row plots a single prompt's per-head initial-frame attention score. The same vertical stripes are coherent across all 50 rows, meaning the same subset of heads attends strongly to the initial frame regardless of prompt. (b) Each row plots the per-head cross-prompt mean attention score at one of the four diffusion timesteps. The four rows display the same vertical stripes, indicating timestep-invariance as well.}
\vspace{-1.5em}
    \label{fig:distri_prompt_step}
\end{figure}

\newparagraph{Stability of the Head Partition.} 
We find the partition pattern to be stable across inputs and training stages. \cref{fig:distri_prompt_step} shows that a largely overlapping subset of heads consistently exhibits elevated initial-frame attention across diverse prompts and diffusion timesteps. The partition is also already present in the ODE-initialization model and the anchor-head locations remain stable after distillation (\eg, comparing Self-Forcing and its ODE-initialization in \cref{fig:temporal_attention}). Notably, a similar partition emerges in the bidirectional teacher backbone (\cref{sec:head_emergence}), suggesting the structure is rooted in pretraining rather than the AR distillation pipeline. Together, these observations indicate that the anchor/working partition is a structural property of the pretrained model rather than an artifact of any particular distillation stage.

\vspace{-0.5em}

\subsection{Motion Collapse from Universal Frame Sink}
\label{sec:collapse}
\vspace{-0.5em}
The above analysis shows that initial frames stabilize global consistency sparsely through the \anchorheads{}. In current sink-based AR-VDMs, however, the initial frames are exposed to every head via the shared KV context $[\textcolor{anchorred}{\bK_{\mathrm{init}}};\, \textcolor{workingblue}{\bK_{\mathrm{win}}};\, \bK_{\mathrm{cur}}]$, $[\textcolor{anchorred}{\bV_{\mathrm{init}}};\, \textcolor{workingblue}{\bV_{\mathrm{win}}};\, \bV_{\mathrm{cur}}]$, including the majority of \workingheads{} that focus on local dynamics. Combined with the mode-seeking DMD objective~\cite{dmd,dmd2}, this universal conditioning offers the student an easy shortcut toward high-fidelity static modes. The initial frames provide a clean reference with minimal exposure bias, unlike the sliding window, which accumulates self-rollout error. Static or near-static outputs that closely track this reference, therefore, tend to score higher on consistency than more dynamic outputs (\cref{fig:qualitative,tab:main}). Such static modes also remain probable under the teacher distribution, since pretraining video datasets usually contain a substantial fraction of static scenes. With universal access to $\textcolor{anchorred}{\bK_{\mathrm{init}}}$ and $\textcolor{anchorred}{\bV_{\mathrm{init}}}$, every head---including the \workingheads{} normally responsible for local dynamics---can exploit this shortcut, leading the student to over-condition on the initial frames at the cost of motion.

\section{The Disen-Forcing Method}
\label{sec:method}
\vspace{-0.7em}
Building on the above analysis, frame sink rests on a universal attention sink premise that does not hold in AR-VDMs and overlooks the observed head specialization. We therefore propose \textbf{Disen-Forcing}, a distillation-time plug-in for AR-VDMs that respects this specialization during distillation. Given an ODE-initialization model (\cref{sec:preliminary}), we first perform a one-time identification of anchor heads, then distill with two complementary components: \textit{disentangled context routing} (\cref{sec:contextsep}) that gives each head type its functional context, and \textit{motion-disentangling regularization} (\cref{sec:reg}) that prevents anchor heads from inducing motion collapse.

\vspace{-0.5em}

\subsection{Disentangled Context Routing}
\label{sec:contextsep}
\vspace{-0.5em}
The motion collapse issue reveals an over-conditioning of the frame sink heuristic on initial-frame context. Inspired by the observation of specialized head types in \cref{sec:uncover}, we propose to assign each head type a dedicated context that matches its functional role, shielding \workingheads{} from the static initial-frame reference and thereby eliminating the over-reliance at its source.

\newparagraph{Identifying Anchor Heads.}
\cref{fig:distri_prompt_step} shows that the anchor/working partition is a stable, input-independent property of the pretrained backbone. We therefore identify the set of anchor heads $\cA$ by computing the average initial-frame attention score $s_{l,h} = \mathbb{E}_{\text{prompt},t,i}\bigl[A^{i,1}_{l,h}\bigr]
$ of the $h$-th head at layer-$l$, averaged over prompts, 
diffusion steps $t$, and AR steps $i$ and thresholding at the $p$-th percentile $\tau_p$ across all heads:
$\cA_l = \bigl\{h : s_{l,h} > \tau_p\bigr\}$ and $\cA = \bigl\{(l,h) : h \in \cA_l\bigr\}$. The set $\cA$ is computed once using each base model's ODE-initialization checkpoint~\cite{selfforcing,causalforcing} before training and kept fixed throughout distillation, which is supported by the consistent attention pattern observed across ODE-initialization and post-distillation baselines in \cref{fig:temporal_attention}.

\newparagraph{Context Routing.}
With the anchor head set $\cA$ identified, we route KV context per head type. Let $l$ index layers and $h \in \{1, \ldots, H\}$ index the attention heads within each layer. We route KV context per head as:
\begin{equation}
    \bK^{l,h}_{\mathrm{ctx}} = \begin{cases}
        [\textcolor{anchorred}{\bK^{l,h}_{\mathrm{init}}};\, \bK^{l,h}_{\mathrm{cur}}] & \text{if } h \in \cA_l, \\
        [\textcolor{workingblue}{\bK^{l,h}_{\mathrm{win}}};\, \bK^{l,h}_{\mathrm{cur}}] & \text{otherwise,}
    \end{cases}
    \label{eq:routing}
\end{equation}
and analogously for $\bV^{l,h}_{\mathrm{ctx}}$.
\workingheads{} attend only to the local sliding window and the current frame---the temporally local context they need to construct motion. \anchorheads{} attend only to the initial frame and the current frame, retaining their clean global reference without unnecessary local context that may carry cumulative error. Importantly, both head types retain the same context budget---substituting $\textcolor{anchorred}{\bK_{\mathrm{init}}}$ for $\textcolor{workingblue}{\bK_{\mathrm{win}}}$ rather than expanding or truncating the context window.

\subsection{Motion-Disentangling Regularization}
\label{sec:reg}

Although context routing ensures that \workingheads{} attend exclusively to temporally local context, freeing their motion representations from initial-frame interference, \anchorheads{}---though sparse---retain a path to initial frames and can still leak global semantics into motion construction. Routing alone cannot fully close this leakage: it constrains only which context each head reads, not which aspect of the generated video each head shapes. We therefore impose an invariance constraint directly on the distillation process: the local motion construction must remain invariant under intervention on \anchorheads{}' access to initial frames.

\newparagraph{Motion Proxy.} Enforcing this invariance requires a measurable proxy for motion, which we extract from the model's internal representations. Analysis of video diffusion transformers shows that cross-frame query--key similarity captures temporal correspondence in selected layers and denoising timesteps, supporting zero-shot point tracking~\cite{difftrack}. Query--key matching has also been used for temporal cache compression~\cite{tempcache}. We therefore adopt the cross-frame self-attention map $\bA^{i,j} = \mathrm{Softmax}(\bQ^i (\bK^j)^\top / \sqrt{d})$ as our motion proxy. Concretely, $\bA_l^{i,i-1}$ at layer $l$ is, for each spatial token of frame $i$, a soft correspondence over the positions of frame $i{-}1$, approximating the local motion field at that layer. The proxy is differentiable and computed from the queries and keys in the forward pass.

Let $\bA_l^{i,i-1}$ denote the motion proxy from the standard forward pass with disentangled routing (\cref{eq:routing}). We run a counterfactual pass with the same model and input, replacing \anchorheads{}' routed context with $[\textcolor{workingblue}{\bK_{\mathrm{win}}};\,\bK_{\mathrm{cur}}]$ (and analogously for $\bV$), and denote its proxy by $\hat{\bA}_l^{i,i-1}$. We penalize the discrepancy between the two proxies using KL divergence over layers $\mathcal{S}$ that contain no \anchorheads{}:
\begin{equation}
    \mathcal{L}_{\mathrm{reg}} = \frac{1}{|\mathcal{S}|}
    \sum_{l \in \mathcal{S}}
    \frac{1}{N} \sum_{i=1}^{N}
    \operatorname{KL}\bigl(\operatorname{sg}(\hat{\bA}_l^{i,i-1}) \,\|\, \bA_l^{i,i-1}\bigr).
    \label{eq:reg_loss}
\end{equation}
Here $\operatorname{sg}(\cdot)$ denotes stop-gradient. Minimizing $\mathcal{L}_{\mathrm{reg}}$ encourages the motion proxy in non-anchor layers to be invariant to whether anchor heads have access to the initial frames, thereby disentangling semantic anchoring from motion construction. The full training objective is $\mathcal{L} = \mathcal{L}_{\mathrm{DMD}} + \lambda_{\mathrm{reg}}\mathcal{L}_{\mathrm{reg}}$.

\begin{figure*}[ht]
\centering
\includegraphics[width=\linewidth]{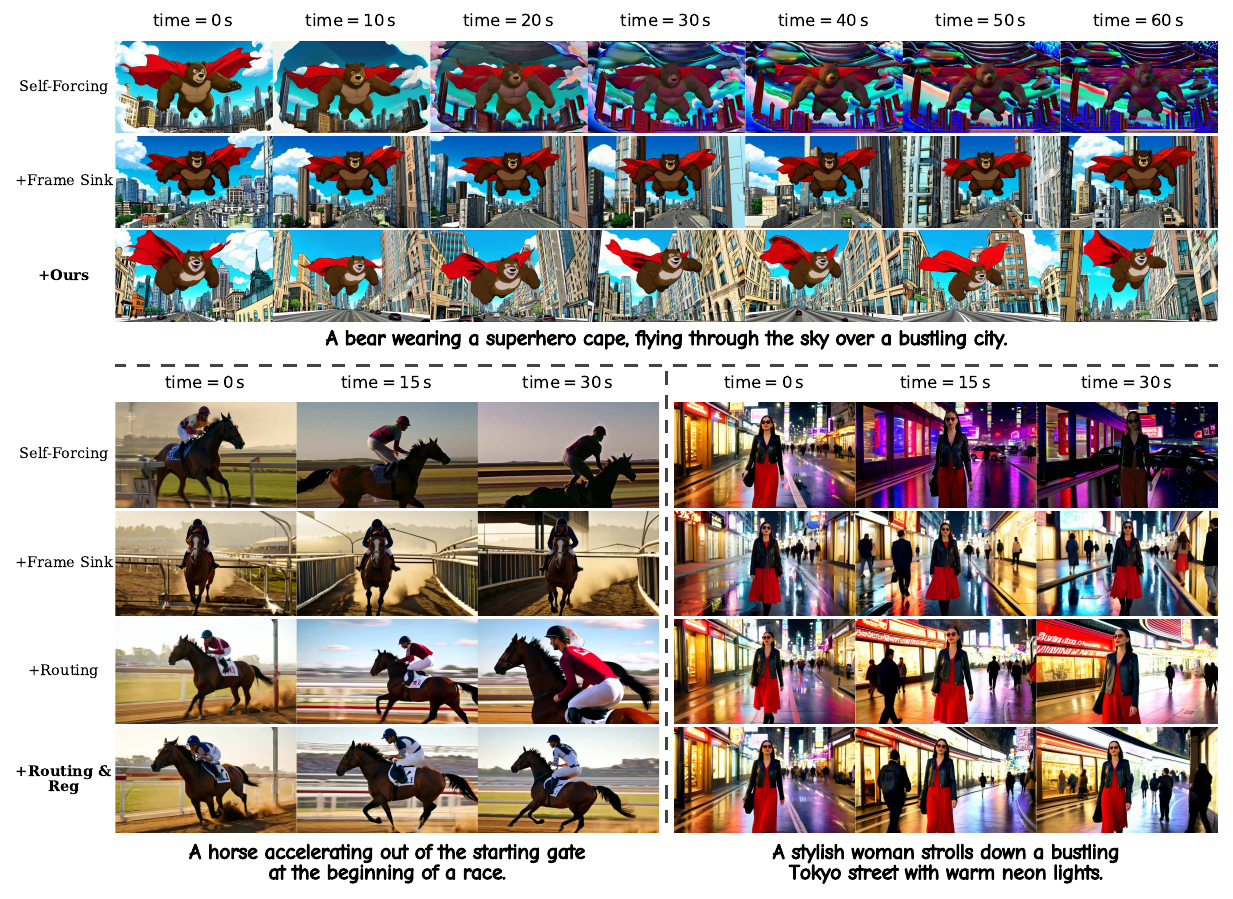}
\caption{\textbf{Qualitative comparison and ablation.}
Self-Forcing suffers from temporal drift once the inference rollout exceeds the training horizon; Frame Sink avoids temporal drift but yields low-dynamic, conservative motion.
Ours (Disen-Forcing) is both drift-free and temporally dynamic.
The bottom block isolates the contributions of our two components: context routing and $\mathcal{L}_{\mathrm{reg}}$.}
\vspace{-1em}
\label{fig:qualitative}
\end{figure*} 
\vspace{-0.5em}
\section{Experiments}
\label{sec:experiments}
\vspace{-0.5em}
\newparagraph{Implementation details.} We implement Disen-Forcing as a distillation-time plug-in on Self-Forcing~\cite{selfforcing} and Causal-Forcing~\cite{causalforcing}. Both base models are four-step causal AR-VDMs based on Wan2.1-T2V-1.3B, with a 5-second training horizon and an 832$\times$480 resolution. We finetune each base model from its own ODE-initialization checkpoint with a KV cache size of 9 latent frames---3 initial, 3 sliding-window, and 3 current frames. Each AR rollout step denoises a chunk of 3 latent frames, and the 5-second training horizon corresponds to 7 such rollout steps. Distillation runs on VidProM~\cite{vidprom}. We use a batch size of 64, except for Self-Forcing + Disen-Forcing and the ablations, which use 32. The anchor head set $\cA$ is identified using all 1003 MovieGenBench~\cite{moviegen} prompts by thresholding at the 80th percentile. Its stability across prompts and sample sizes is evaluated in \cref{fig:distri_prompt_step,fig:anchor-robustness}. We apply motion-disentangling regularization to layer~26 ($\mathcal{S} = \{26\}$ in \cref{eq:reg_loss}) with weight $\lambda_{\mathrm{reg}} = 0.002$. Layer selection is evaluated in \cref{sec:proxy_layer}. See \cref{sec:impl} for more details.

\newparagraph{Evaluation.} We evaluate on 200 MovieGenBench~\cite{moviegen} prompts using the indices published by Rolling-Forcing~\cite{rollingforcing}. We generate 30-second and 1-minute videos at 16 fps, covering the long-video horizons also studied in prior work~\cite{deepforcing}. We use VBench-Long~\cite{huang2024vbench,huang2026vbenchpp} across six dimensions: subject consistency, background consistency, motion smoothness, dynamic degree, aesthetic quality, and imaging quality. We report all six dimensions on a 0--100 scale and their mean (Total). Following \citet{framepack}, temporal drift is measured as the mean absolute imaging quality gap between the first and last segments ($\Delta^{\mathrm{Q}}_{\mathrm{Drift}}$).
\vspace{-0.5em}
\subsection{Comparison with Existing Models}
\begin{table}[t]
\caption{\textbf{Quantitative comparison on long video generation.}
Disen-Forcing is a distillation-time plug-in applied to Self-Forcing and Causal-Forcing.
$\Delta^{\mathrm{Q}}_{\mathrm{Drift}}$\,($\downarrow$) measures the imaging quality gap between the first and last segments.
Best Total per horizon in \textbf{bold}.}
\label{tab:main}
\centering
\setlength{\tabcolsep}{5.0pt}
\renewcommand{\arraystretch}{1.08}
\resizebox{\textwidth}{!}{%
\begin{tabular}{@{}l ccccccc c@{}}
\toprule
& \multicolumn{7}{c}{VBench-Long Scores\,$\uparrow$}
& \multirow{2}{*}{$\Delta^{\mathrm{Q}}_{\mathrm{Drift}}$\,$\downarrow$} \\
\cmidrule(lr){2-8}
Model
& Total
& Subject
& Background
& Smooth
& Dynamic
& Aesthetic
& Imaging
& \\

\midrule
\multicolumn{9}{c}{\cellcolor{groupgray!300}\textit{30-second generation ($\sim$500 frames)}} \\
\midrule
\rowcolor{groupgray}
\multicolumn{9}{@{}l}{\textit{Diffusion Forcing}} \\
SkyReels-V2~\cite{skyreels}
& 76.64 & 96.89 & 96.98 & 99.14 & 45.50 & 60.04 & 61.31 & 5.65 \\
\midrule
\rowcolor{groupgray}
\multicolumn{9}{@{}l}{\textit{Distilled Causal}} \\
CausVid~\cite{causvid}
& 78.44 & 97.83 & 96.72 & 98.15 & 52.59 & 59.33 & 66.01 & 6.73 \\
Rolling-Forcing~\cite{rollingforcing}
& 76.16 & 97.50 & 96.37 & 98.67 & 33.53 & 60.88 & 70.01 & 3.70 \\
LongLive~\cite{longlive}
& 77.51 & 97.40 & 96.37 & 98.77 & 42.50 & 61.68 & 68.34 & 3.74 \\
Deep-Forcing~\cite{deepforcing}
& 78.58 & 95.40 & 94.81 & 97.96 & 58.09 & 58.93 & 66.26 & 5.80 \\
\rowcolor{sflight}
Self-Forcing~\cite{selfforcing}
& 76.37 & 96.49 & 95.97 & 98.33 & 40.44 & 59.09 & 67.87 & 7.23 \\
\rowcolor{sfmid}
\quad + Frame Sink
& 76.93 & 97.91 & 96.73 & 98.94 & 35.66 & 62.37 & 69.94 & 3.40 \\
\rowcolor{sfdark}
\quad \textbf{+Disen-Forcing}
& \textbf{82.38} & 96.52 & 95.50 & 98.13 & 73.12 & 60.95 & 70.05 & 4.03 \\
\rowcolor{cflight}
Causal-Forcing~\cite{causalforcing}
& 79.65 & 95.53 & 95.13 & 97.30 & 68.28 & 55.92 & 65.75 & 9.38 \\
\rowcolor{cfmid}
\quad + Frame Sink
& 74.12 & 98.07 & 96.91 & 98.91 & 17.75 & 61.64 & 71.45 & 3.25 \\
\rowcolor{cfdark}
\quad \textbf{+Disen-Forcing}
& 81.57 & 95.93 & 95.34 & 98.00 & 69.72 & 60.66 & 69.77 & 4.83 \\
\midrule
\multicolumn{9}{c}{\cellcolor{groupgray!300}\textit{1-minute generation ($\sim$1000 frames)}} \\
\midrule

\rowcolor{groupgray}
\multicolumn{9}{@{}l}{\textit{Distilled Causal}} \\
CausVid
& 78.01 & 98.01 & 96.91 & 98.24 & 50.33 & 59.13 & 65.48 & 7.51 \\
Rolling-Forcing
& 75.88 & 97.39 & 96.26 & 98.69 & 32.38 & 60.17 & 70.38 & 4.18 \\
LongLive
& 77.67 & 97.01 & 96.15 & 98.73 & 43.97 & 61.69 & 68.46 & 3.70 \\
Deep-Forcing
& 79.35 & 96.30 & 95.47 & 98.18 & 58.90 & 59.46 & 67.77 & 5.04 \\
\rowcolor{sflight}
Self-Forcing
& 74.75 & 96.82 & 96.34 & 98.38 & 34.70 & 56.55 & 65.68 & 12.73 \\
\rowcolor{sfmid}
\quad + Frame Sink
& 76.88 & 97.79 & 96.58 & 98.91 & 35.83 & 61.81 & 70.34 & 3.10 \\
\rowcolor{sfdark}
\quad \textbf{+Disen-Forcing}
& \textbf{82.34} & 96.39 & 95.65 & 98.06 & 73.95 & 59.97 & 70.01 & 4.04 \\
\rowcolor{cflight}
Causal-Forcing
& 77.71 & 95.35 & 95.25 & 96.97 & 64.91 & 51.22 & 62.57 & 12.96 \\
\rowcolor{cfmid}
\quad + Frame Sink
& 73.68 & 98.01 & 96.81 & 98.90 & 16.06 & 60.79 & 71.51 & 3.66 \\
\rowcolor{cfdark}
\quad \textbf{+Disen-Forcing}
& 81.26 & 95.76 & 95.26 & 98.00 & 69.64 & 59.92 & 68.96 & 5.10 \\
\bottomrule
\end{tabular}%
}
\end{table}
\newparagraph{Baselines.} Our primary baselines are the Frame Sink variants of Self-Forcing and Causal-Forcing, each distilled from the same ODE checkpoint as its Disen-Forcing counterpart under matched compute (identical KV cache size and iterations). This isolates the contribution of Disen-Forcing beyond the frame sink alone. We further compare against three sliding-window AR-VDMs (CausVid~\cite{causvid}, Self-Forcing~\cite{selfforcing}, Causal-Forcing~\cite{causalforcing}), three sink-based methods (LongLive~\cite{longlive}, Rolling-Forcing~\cite{rollingforcing}, Deep-Forcing~\cite{deepforcing}), and the diffusion-forcing model SkyReels-V2~\cite{skyreels}. In our evaluation, all baselines use $4$ denoising steps except Rolling-Forcing ($5$) and SkyReels-V2 ($30$, undistilled).

\newparagraph{Results.} \cref{tab:main} reveals two failure modes of existing models. The base models lose early-frame context as the cache rolls forward and accumulate severe drift, while their Frame Sink variants recover consistency but suppress motion. Disen-Forcing improves both motion dynamics and long-range consistency over both base models. On Self-Forcing, it increases Dynamic Degree from Frame Sink's $35.66$/$35.83$ to $73.12$/$73.95$ at the 30-second/1-minute horizons, while reducing $\Delta^{\mathrm{Q}}_{\mathrm{Drift}}$ from the base model's $7.23$/$12.73$ to $4.03$/$4.04$. On Causal-Forcing, it similarly increases Dynamic Degree from Frame Sink's $17.75$/$16.06$ to $69.72$/$69.64$ and reduces $\Delta^{\mathrm{Q}}_{\mathrm{Drift}}$ from the base model's $9.38$/$12.96$ to $4.83$/$5.10$. We find that the base models' low Dynamic Degree at long horizons is largely caused by accumulated drift, which explains why Self-Forcing scores slightly below its Frame Sink variant at 1 minute. At the 5s training horizon, without frame-length extrapolation, Frame Sink reduces Dynamic Degree from $65.00$ to $37.00$ for only marginal consistency gains, whereas Disen-Forcing raises it to $74.00$ (\cref{sec:5s_results}), confirming that universal sink exposure harms motion under DMD. As the rule-based consistency and drift metrics favor near-static videos, we also use a VLM-based human-preference judge, which prefers Disen-Forcing over Frame Sink across all five dimensions on both base models, including visual quality and temporal consistency (\cref{sec:videoscore}).
\subsection{Ablation Studies}
We conduct ablations to assess the contribution of our two core components---context routing and motion-disentangling regularization---starting from the Self-Forcing+Frame Sink baseline, as summarized in \cref{tab:ablation} and visualized qualitatively in \cref{fig:qualitative}. We then examine the anchor-head threshold $\tau_p$ and compare the identified anchor-head set $\cA$ with layer-matched and globally random sets to assess the importance of head selection. Finally, we study the regularization-layer selection $\mathcal{S}$ by comparing individual candidate layers with their joint regularization, and vary the regularization weight $\lambda_{\mathrm{reg}}$ to examine its effect on temporal dynamics and consistency. Full per-dimension results and additional ablations are reported in Appendix~\ref{sec:full_ablation}.

\newparagraph{Context Routing.} Adding context routing alone to the baseline lifts dynamic degree from $34.69$ to $52.56$, indicating that exposing only anchor heads to the initial frames---rather than every head as in frame sink---already mitigates the over-conditioning shortcut and recovers a substantial portion of motion. $\Delta^{\mathrm{Q}}_{\mathrm{Drift}}$ simultaneously decreases from $4.40$ to $4.08$, indicating that routing recovers motion without sacrificing long-range consistency.

\newparagraph{Impact of $\tau_p$.} Lower $\tau_p$ ($p{=}70$, more anchors) over-restricts motion ($48.22$ dynamic degree), while higher $\tau_p$ ($p{=}90$, fewer anchors) increases $\Delta^{\mathrm{Q}}_{\mathrm{Drift}}$ ($4.77$). We adopt $p{=}80$ to balance the two. Visual inspection of the head-wise attention grids in \cref{fig:head_full_grid} confirms that $p{=}80$ gives an accurate anchor/working partition.

\begin{wraptable}[34]{r}{0.58\textwidth}
\vspace{-16pt}
\captionsetup{font=small}
\caption{\textbf{Ablation study on main components.} We progressively add disentangled routing and motion-disentangling regularization to the Self-Forcing + Frame Sink baseline. We also ablate the anchor-head threshold $\tau_p$, regularization weight $\lambda_{\mathrm{reg}}$, regularization layer $\mathcal{S}$, and anchor-head set $\cA$ selection strategies. The component and threshold groups use four-layer regularization; other groups use layer~26 unless the regularization layer is varied. Rows are comparable within each group.}
\label{tab:ablation}
\centering\footnotesize
\setlength{\tabcolsep}{2.5pt}
\renewcommand{\arraystretch}{1.15}
\begin{adjustbox}{max width=\linewidth}
\begin{tabular}{@{}l ccccc@{}}
\toprule
& \multicolumn{4}{c}{VBench-Long Scores\,$\uparrow$} & \multirow{2}{*}{$\Delta^{\mathrm{Q}}_{\mathrm{Drift}}$\,$\downarrow$} \\
\cmidrule(lr){2-5}
Setting & Subject & Smooth & Dynamic & Imaging & \\
\midrule
\rowcolor{groupgray}
\multicolumn{6}{@{}l}{\textit{Context routing and regularization}} \\
Self-Forcing + Frame Sink & 97.25 & 98.33 & 34.69 & 70.76 & 4.40 \\
\quad + Routing & 96.93 & 98.48 & 52.56 & 69.82 & 4.08 \\
\quad + Routing \& $\mathcal{L}_{\mathrm{reg}}$ & 95.96 & 97.36 & 77.19 & 68.97 & 4.39 \\
\midrule
\rowcolor{groupgray}
\multicolumn{6}{@{}l}{\textit{Impact of $\tau_p$}} \\
$p{=}70$ & 97.70 & 98.59 & 48.22 & 70.90 & 3.93 \\
$p{=}80$ & 95.96 & 97.36 & 77.19 & 68.97 & 4.39 \\
$p{=}90$ & 95.64 & 97.28 & 83.47 & 68.57 & 4.77 \\
\midrule
\rowcolor{groupgray}
\multicolumn{6}{@{}l}{\textit{Anchor-head set $\cA$ selection}} \\
Identified & 96.52 & 98.13 & 73.12 & 70.05 & 4.03 \\
Layer-matched random & 97.45 & 98.47 & 51.16 & 70.08 & 4.51 \\
Globally random & 98.15 & 98.88 & 34.69 & 70.67 & 3.25 \\
\midrule
\rowcolor{groupgray}
\multicolumn{6}{@{}l}{\textit{Regularization layer $\mathcal{S}$ selection}} \\
Layer~20 & 96.20 & 97.85 & 75.66 & 69.59 & 5.01 \\
Layer~26 & 96.52 & 98.13 & 73.12 & 70.05 & 4.03 \\
Layer~27 & 96.68 & 97.85 & 65.56 & 69.69 & 4.26 \\
Layer~28 & 96.48 & 97.95 & 64.81 & 69.06 & 4.18 \\
Layers 20, 26, 27, 28 & 95.78 & 97.43 & 78.38 & 68.92 & 4.83 \\
\midrule
\rowcolor{groupgray}
\multicolumn{6}{@{}l}{\textit{Impact of $\lambda_{\mathrm{reg}}$}} \\
$\lambda_{\mathrm{reg}} = 0.001$ & 97.07 & 98.31 & 58.47 & 70.21 & 4.72 \\
$\lambda_{\mathrm{reg}} = 0.002$ & 96.52 & 98.13 & 73.12 & 70.05 & 4.03 \\
$\lambda_{\mathrm{reg}} = 0.005$ & 96.24 & 97.92 & 75.97 & 68.90 & 4.47 \\
\bottomrule
\end{tabular}
\end{adjustbox}
\end{wraptable}
\newparagraph{Anchor-head set $\cA$ selection.} Replacing the identified anchor heads with equally sized random sets, keeping regularization fixed at layer~26, reduces Dynamic Degree from $73.12$ to $51.16$ for layer-matched sampling and $34.69$ for global sampling. Layer-matched random routing also increases $\Delta^{\mathrm{Q}}_{\mathrm{Drift}}$ from $4.03$ to $4.51$, supporting the importance of head identity beyond layer placement. Global random routing yields a lower $\Delta^{\mathrm{Q}}_{\mathrm{Drift}}$ ($3.25$), but also a lower Dynamic Degree.

\newparagraph{Impact of $\mathcal{L}_{\mathrm{reg}}$.}
Adding $\mathcal{L}_{\mathrm{reg}}$ further lifts dynamic degree from $52.56$ to $77.19$,
confirming that routing alone does not fully prevent the leakage of initial-frame content into motion construction.
$\mathcal{L}_{\mathrm{reg}}$ explicitly encourages invariance of motion construction under anchor-head intervention, suppressing this leakage and strengthening the separation between motion and global semantic anchoring.

\newparagraph{Motion-proxy layer.}
We select layers 20, 26, 27, and 28 as candidate regularization layers because they contain no identified anchor heads. We probe these layers by warping a reference frame along attention-derived correspondences and measuring reconstruction error relative to an identity warp (\cref{fig:proxy-selection}; Appendix~\ref{sec:proxy_layer}). Layer~26 has the lowest reconstruction error. Regularizing layer~26 also yields the lowest $\Delta^{\mathrm{Q}}_{\mathrm{Drift}}$ ($4.03$), supporting this layer choice. The four-layer variant attains a higher Dynamic Degree ($78.38$ versus $73.12$), at the cost of a higher $\Delta^{\mathrm{Q}}_{\mathrm{Drift}}$.

\newparagraph{Impact of $\lambda_{\mathrm{reg}}$.}
Increasing $\lambda_{\mathrm{reg}}$ from $0.001$ to $0.002$ raises Dynamic Degree from $58.47$ to $73.12$ and reduces $\Delta^{\mathrm{Q}}_{\mathrm{Drift}}$ from $4.72$ to $4.03$. Further increasing $\lambda_{\mathrm{reg}}$ to $0.005$ yields a smaller gain in Dynamic Degree ($75.97$), while increasing $\Delta^{\mathrm{Q}}_{\mathrm{Drift}}$ to $4.47$. We use $\lambda_{\mathrm{reg}}{=}0.002$ to balance the two.
\vspace{-1em}
\section{Conclusion}
\vspace{-1em}
In this work, we revisit the anti-drift role of initial frames in AR-VDMs and find that the implicit assumption underlying existing methods---that initial frames serve as a universal attention sink---does not hold. Instead, anti-drift is carried by a sparse, stable subset of \emph{anchor heads}, separate from majority \emph{working heads} that attend locally. Building upon this observation, we propose \textbf{Disen-Forcing}, a distillation-time plug-in with two components: \emph{disentangled context routing} restricts initial-frame access to anchor heads, and \emph{motion-disentangling regularization} enforces motion construction to remain invariant under initial-frame conditioning. Extensive experiments show that Disen-Forcing improves both motion dynamics and long-range consistency over the base AR-VDMs.

\acksection
\label{sec:acknowledgement}
We sincerely thank Zhaoyuan Yang for helpful methodological discussions and April Li for constructive feedback on the presentation of this paper. We also sincerely thank Dongseok Shim and Michael Yeung for their valuable feedback on our earlier work in this direction. Zhengyang Yu is supported by the ANU Postgraduate Research Scholarship and the ANU HDR Fee Remission Merit Scholarship.

\newpage

\appendix
\setcounter{section}{0}
\setcounter{figure}{0}
\setcounter{table}{0}
\renewcommand{\thesection}{\Alph{section}}
\renewcommand{\thesubsection}{\Alph{section}.\arabic{subsection}}
\renewcommand{\thetable}{A\arabic{table}}
\renewcommand{\thefigure}{A\arabic{figure}}

\section{Implementation Details}
\label{sec:impl}

\subsection{Training Configurations}

\paragraph{Training setup.} We use AdamW with $(\beta_1, \beta_2) = (0.0, 0.999)$ for both the generator and the critic, with learning rates $2{\times}10^{-6}$ and $4{\times}10^{-7}$ respectively. The critic and generator are updated in a 5:1 ratio (5 critic updates per generator update). We apply an exponential moving average (EMA) on the generator weights with decay $0.99$, starting at effective step 200. Models are sharded across 8 NVIDIA H100 GPUs using FSDP with hybrid-full sharding and size-based wrapping. We enable gradient checkpointing during training. A full training run of 600 effective steps takes approximately 14 hours.

\paragraph{RoPE configuration.} We store keys without RoPE in the KV cache and apply RoPE when using the cached keys to compute attention. Initial frames are positioned immediately before the local sliding window in temporal position space, following Infinite-Forcing~\cite{infiniteforcing} and Rolling-Forcing~\cite{rollingforcing}.

\subsection{More Details on Motion-Disentangling Regularization}
\label{sec:two_pass}
\begin{figure}[t]
\centering
\begin{tikzpicture}[x=1cm,y=1cm,>=Stealth,
 box/.style={draw=black,line width=1pt,rounded corners=1.5pt,inner sep=3pt},
 fwd/.style={box,font=\footnotesize,minimum width=2cm,minimum height=.7cm},
 card/.style={box,font=\footnotesize,align=center,text width=3.65cm,
   execute at begin node={\hyphenpenalty=10000}},
 head/.style={font=\footnotesize,anchor=base},
 arr/.style={-{Stealth[length=3.5pt,width=3.5pt]},draw=black,line width=1pt}]
\draw[dashed,draw=black!40,line width=.45pt] (-.97,0) rectangle (5.95,-2.7);
\draw[dashed,draw=black!40,line width=.45pt] (6.05,0) rectangle (12.97,-2.7);
\node[font=\footnotesize,anchor=base west] at (-.7,-2.45) {Standard pass};
\node[font=\footnotesize,anchor=base east] at (12.7,-2.45) {Counterfactual pass};
\node[box,font=\footnotesize,align=center,minimum width=2cm,minimum height=.75cm] (in) at (6,.85) {Input\\$\bx_t,t,c$};
\node[fwd,align=center,text width=2.5cm,inner sep=1.5pt,minimum height=.85cm] (fw) at (4.6,-.85)
  {Model forward\\[-1pt]{\textcolor{black!65}{\mbox{(disentangled routing)}}}};
\node[fwd,align=center,text width=2.5cm,inner sep=1.5pt,minimum height=.85cm] (cffw) at (7.4,-.85) {Model forward\\[-1pt]{\textcolor{black!65}{(stop gradient)}}};
\node[card] (std) at (1.15,-.85)
  {{\footnotesize Context caching}\\[3pt]
   $\bz_{\mathrm{ctx}}=[\bz_{\mathrm{init}};\bz_{\mathrm{local}}]$\\[3pt]
   $\mathrm{KV}\leftarrow\mathrm{Cache}(\bz_{\mathrm{ctx}})$};
\node[card] (cf) at (10.85,-.85)
  {{\footnotesize Context caching}\\[3pt]
   $\tilde{\bz}_{\mathrm{ctx}}\leftarrow\text{VAE re-encode}(\bz_{\mathrm{local}})$\\[3pt]
   $\widetilde{\mathrm{KV}}\leftarrow\mathrm{Cache}(\tilde{\bz}_{\mathrm{ctx}})$};
\node[fwd,minimum width=1.6cm] (a)    at (4.6,-2.0) {$\bA_l^{i,i-1}$};
\node[fwd,minimum width=1.6cm] (ahat) at (7.4,-2.0) {$\hat{\bA}_l^{i,i-1}$};
\node[box,font=\footnotesize,minimum width=3.4cm,minimum height=.55cm] (kl) at (6,-3.35) {KL divergence};
\draw[arr] (in.west) -| (fw.north);
\draw[arr] (in.east) -| (cffw.north);
\draw[arr] (3.09,-.85) -- (3.29,-.85);
\draw[arr] (8.91,-.85) -- (8.71,-.85);
\draw[arr] (fw.south)   -- (a.north);
\draw[arr] (cffw.south) -- (ahat.north);
\draw[arr] (a.south)    -- (a.south |- kl.north);
\draw[arr] (ahat.south) -- (ahat.south |- kl.north);
\end{tikzpicture}
\caption{\textbf{Computation flow of the motion-disentangling regularization.} The standard and counterfactual passes share the same input and model parameters but use different context caches. Adjacent-frame attention maps from the two passes are aligned by KL divergence, with the counterfactual map serving as a stop-gradient target.}
\label{fig:two_pass}
\end{figure}
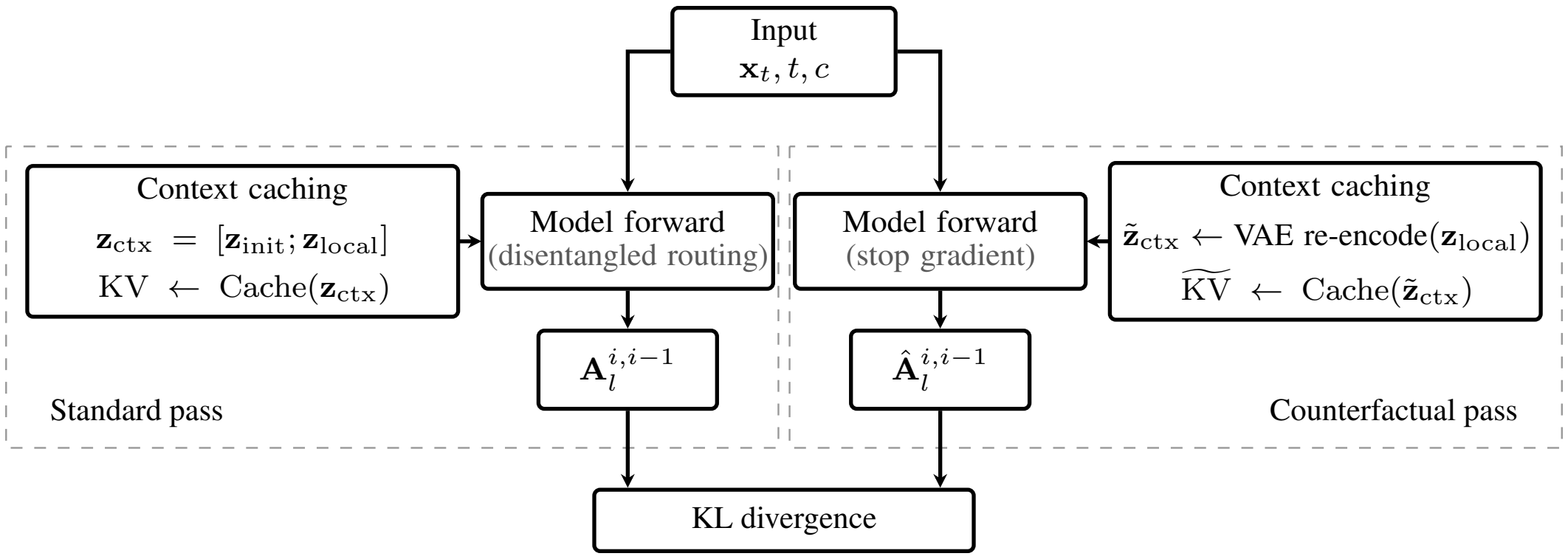

\paragraph{Regularization computation.} \cref{fig:two_pass} summarizes the computation. The motion-disentangling regularization compares cross-frame attention maps from two passes: the standard forward (with disentangled routing) and the counterfactual pass (with \anchorheads{}' KV replaced). We use forward KL divergence $\mathrm{KL}(\operatorname{sg}(\hat{\bA})\,\|\,\bA)$, with the counterfactual proxy $\hat{\bA}$ as the stop-gradient target for the standard proxy $\bA$. For each query frame $i$, the cross-frame attention map is computed against the immediately preceding frame $i{-}1$ (\emph{adjacent matching}); its effect is evaluated in \cref{tab:ablation_full}. Without adjacent matching, all query frames instead match the oldest frame in the sliding window.

\paragraph{VAE re-encoding for the counterfactual pass.} Wan-VAE~\cite{wan2025wan} uses causal temporal convolutions with feature caching: the initial chunk has no preceding feature context, unlike later chunks. This creates a distribution difference between initial- and later-chunk latents. Since \anchorheads{} are trained to attend to these initial-chunk tokens, directly using sliding-window tokens as the anchor-side keys/values in the counterfactual pass places \anchorheads{} out-of-distribution. To eliminate this token distribution gap, before each counterfactual pass, we VAE-decode the sliding-window context to pixels and re-encode it back to latents \emph{without conditioning on prior context}, mirroring the encoding pathway of the initial chunk. \cref{tab:ablation_full} evaluates the effect of this re-encoding.

\paragraph{Attention-map computation.} The model forward retains FlashAttention~\cite{dao2022flashattention}. We compute the motion proxy separately from the queries and keys using an explicit $\mathrm{Softmax}(\bQ_l^i(\bK_l^{i-1})^\top/\sqrt{d})$ between adjacent generated frames, only at the regularized layers.

\paragraph{Regularized layers.} Layers 2, 3, 20, 26, 27, and 28 contain no identified anchor heads for Self-Forcing. We exclude layers 2 and 3 because regularization at these shallow layers provides only a limited constraint, leaving layers 20, 26, 27, and 28 as the candidates in our layer ablation. The default configuration regularizes layer~26, whereas the component and threshold ablations in \cref{tab:ablation} regularize all four candidate layers with $\lambda_{\mathrm{reg}}=0.01$. The layer probe is reported in \cref{sec:proxy_layer}.

\paragraph{Training cost.} Regularization is computed only at the gradient-carrying denoising step of a generator update. \cref{tab:efficiency} reports the training overhead, including the counterfactual pass, VAE re-encoding, and motion-proxy computation.

\subsection{Anchor Head Identification}
We identify the anchor-head set $\cA$ separately for Self-Forcing and Causal-Forcing on their respective ODE-initialization checkpoints, using the same identification procedure. We use all 1003 prompts from MovieGenBench~\cite{moviegen}; for each prompt, the per-head average initial-frame attention score $s_{l,h}$ is averaged across all AR rollout steps and all 4 diffusion timesteps. The resulting $\{s_{l,h}\}$ is thresholded at the 80th percentile to obtain $\cA$, which is held fixed throughout distillation.

\begin{figure}[h]
    \centering
    \begin{minipage}{0.495\textwidth}
        \centering
        \includegraphics[width=\linewidth]{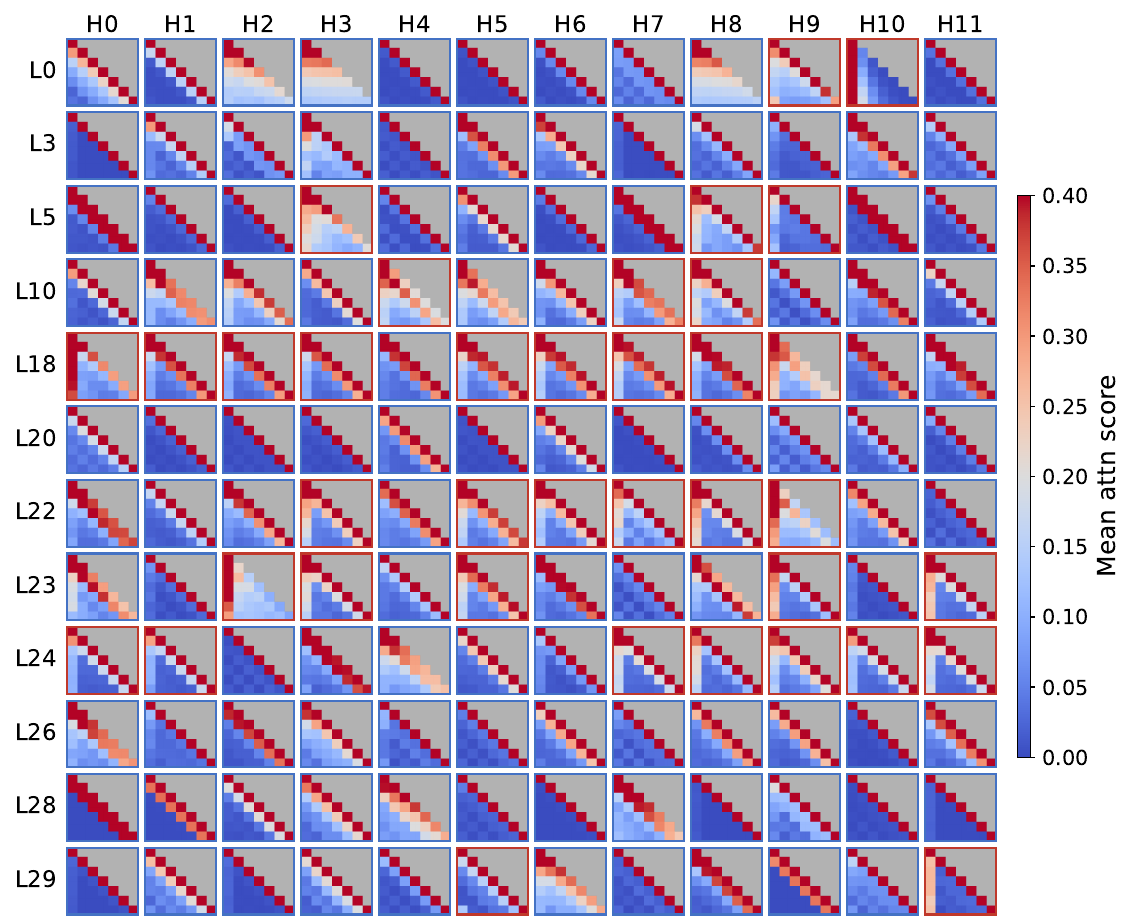}
        \subcaption{Self-Forcing}
    \end{minipage}
    \hfill
    \begin{minipage}{0.495\textwidth}
        \centering
        \includegraphics[width=\linewidth]{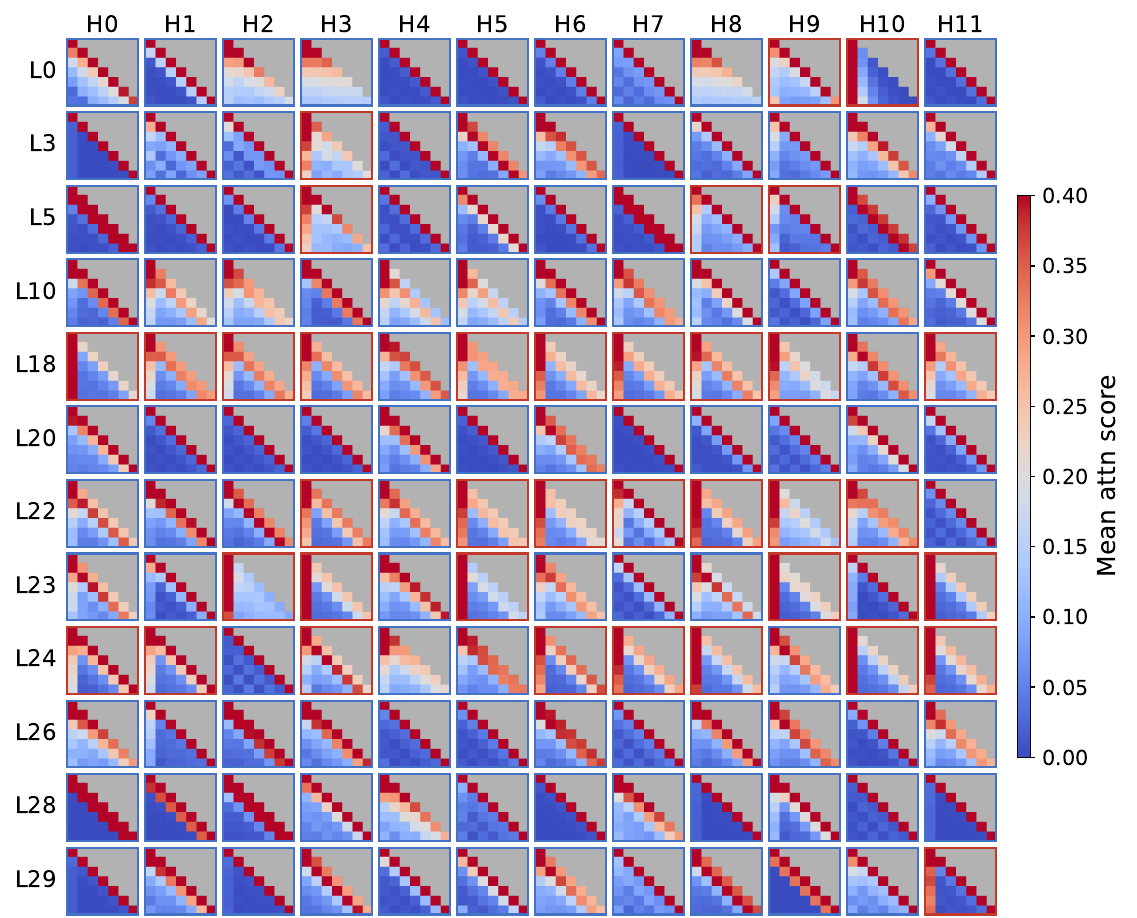}
        \subcaption{Causal-Forcing}
    \end{minipage}
    \caption{\textbf{Representative per-(layer, head) attention map grid.}
    Each cell shows the chunk-by-chunk cross-frame attention pattern of one (layer, head) pair throughout a 7-step AR rollout, averaged over 50 MovieGenBench prompts and all diffusion timesteps.
    \textcolor{anchorred}{\textbf{Red}}-bordered cells are \anchorheads{} (top-20\% by initial-frame attention score, $p=80$); \textcolor{workingblue}{\textbf{blue}}-bordered cells are \workingheads{}.
    The eight heads visualized in \cref{fig:headwise} are sampled from this grid; the representative visualization confirms that the anchor/working partition is sparse, layer-localized (notably layers 18 and 22--24), and consistent across both base models. Zoom in for details.}
    \label{fig:head_full_grid}
\end{figure}

\cref{fig:head_full_grid} shows the representative per-(layer, head) attention map for both base models. For visualization, each cell shows the chunk-by-chunk cross-frame attention pattern of one head averaged over 50 MovieGenBench prompts and all diffusion timesteps; \textcolor{anchorred}{\textbf{red}} borders mark heads selected as \anchorheads{} by thresholding the average initial-frame attention score $s_{l,h}$ at the 80th percentile (\cref{sec:contextsep}).

The visualization grounds three design choices in our identification protocol:
(i)~\textbf{Percentile-thresholding selection.} The vast majority of cells exhibit the canonical local-diagonal pattern of \workingheads{}, with only a small subset showing concentrated initial-frame attention. Our top-20\% percentile threshold aligns with this sparse pattern.
(ii)~\textbf{Layer-localization.} Anchor heads concentrate most strongly in specific layers (most notably layers 18 and 22--24), making the partition structurally identifiable from a small calibration sample.
(iii)~\textbf{Cross-model consistency.} The anchor cluster appears in nearly the same layers for Self-Forcing and Causal-Forcing despite different ODE-initialization checkpoints, suggesting the partition is a structural property of the pretrained Wan2.1 backbone rather than an artifact of either AR training pipeline. Applying the same identification procedure to the distilled Self-Forcing and Causal-Forcing models yields anchor sets with an IoU of $0.78$.

\begin{figure}[h]
    \centering
    \includegraphics[width=\linewidth]{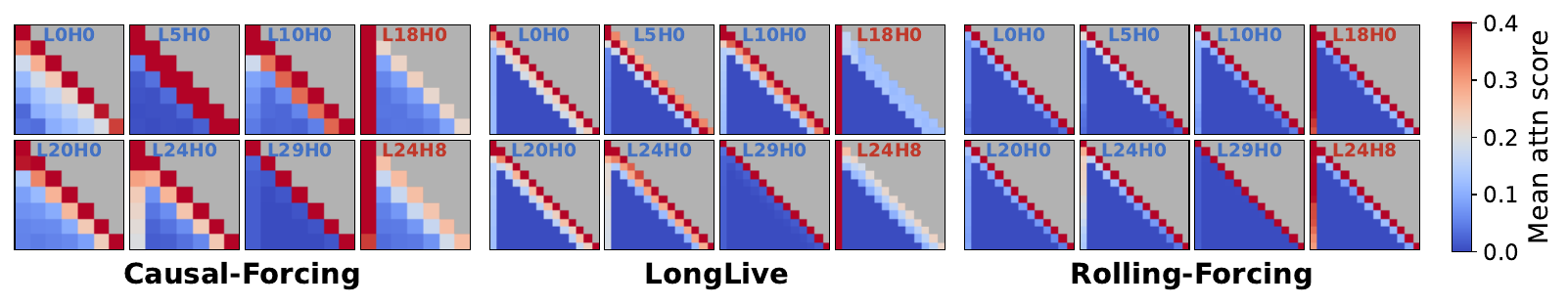}
    \caption{\textbf{Per-head attention distribution for additional baselines.}
    Same head selection and rendering as \cref{fig:headwise}, applied to Causal-Forcing (sliding-window) and the two sink-based methods LongLive and Rolling-Forcing. The same anchor/working partition appears in all three distillation pipelines.}
    \label{fig:headwise_supp}
\end{figure}

\paragraph{Per-head attention on additional baselines.} \cref{fig:headwise_supp} extends \cref{fig:headwise} with the same per-head visualization for the second sliding-window base model Causal-Forcing and two existing sink-based methods, LongLive and Rolling-Forcing. All three exhibit the same anchor/working head partition observed on Self-Forcing, despite the sink-based methods using a 14-chunk rollout (vs.\ 7 in the sliding-window models). This supports our claim that the partition is structural and persists across distillation paradigms.

\paragraph{Sensitivity to prompt count.}

\begin{figure}[h]
  \centering
  \begin{minipage}[c]{0.55\linewidth}
    \centering
    \includegraphics[width=\linewidth]{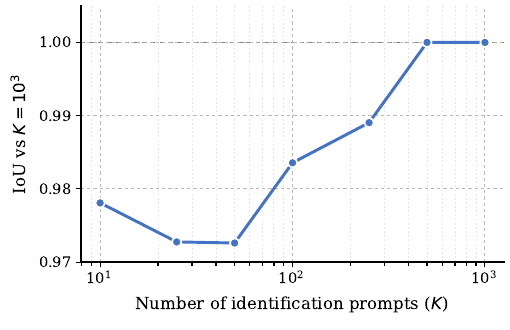}
  \end{minipage}\hfill
  \begin{minipage}[c]{0.42\linewidth}
    \centering
    \begin{tabular}{rcc}
\toprule
$K$ & IoU with $K{=}10^{3}$ & $|\mathcal{A}_K|$ \\
\midrule
10 & $0.978$ & 72 \\
25 & $0.973$ & 72 \\
50 & $0.973$ & 72 \\
100 & $0.984$ & 72 \\
250 & $0.989$ & 72 \\
500 & $1.000$ & 72 \\
1000 & $1.000$ (ref) & 72 \\
\bottomrule
\end{tabular}

  \end{minipage}
  \caption{IoU of the $p_{80}$-thresholded \anchorheads{} set
    $\mathcal{A}_K$ identified from $K$ random prompts versus the reference
    $\mathcal{A}_{10^3}$.  Set size $|\mathcal{A}_K|=72$ at every $K$, out
    of 360 total attention heads.}
  \label{fig:anchor-robustness}
\end{figure}

To test the sensitivity of anchor-head identification to the size of the
identification prompt set, we re-apply $p_{80}$ thresholding on random
subsets of size $K\in\{10, 25, 50, 100, 250, 500\}$ drawn from a
1000-prompt MovieGenBench~\cite{moviegen} pool and measure IoU against the set identified using all 1000 prompts in this pool
(averaged across 5 random draws per $K$).
The recovered \anchorheads{} set is highly stable:
mean $\mathrm{IoU}\geq 0.97$ across all $K$, reaching exact agreement from
$K{=}500$, with a constant set size $|\mathcal{A}|{=}72$ at every $K$
(Fig.~\ref{fig:anchor-robustness}).
The identified set remains stable even with 10 prompts in this experiment.

\subsection{Inference Setup}

At inference time, we use 4-step denoising with timesteps $[1000, 750, 500, 250]$ and timestep shift $5.0$, and use the EMA generator weights for all evaluations. For long-horizon generation, the AR rollout extends autoregressively over the desired horizon (7 chunks for the 5-second training horizon; 42 and 84 chunks for 30-second and 1-minute evaluation, respectively), using the rolling KV cache of the base model.

\subsection{Training Overhead}
We measure the mean forward-and-backward time separately for critic and generator passes. \cref{tab:efficiency} also reports the amortized time under the 1:5 generator-to-critic update schedule.
\begin{table}[h]
\centering
\caption{\textbf{Training overhead.} Critic and generator columns report mean forward-and-backward time per pass. The amortized time is $(5\,t_{\mathrm{critic}}+t_{\mathrm{gen}})/5$ under the 1:5 generator-to-critic update schedule.}
\label{tab:efficiency}
\small
\setlength{\tabcolsep}{5pt}
\begin{tabular}{lcccc}
\toprule
Method & Critic (s) & Generator (s) & Gen:critic & Amortized (s) \\
\midrule
+ Frame Sink   & 7.0 & 16.1 & 1:5 & 10.2 \\
+Ours         & 7.0 & 18.8 & 1:5 & 10.8 ($+5\%$) \\
\bottomrule
\end{tabular}
\end{table}

The additional generator-side cost is amortized over five critic updates per generator update; the measured critic-pass time is unchanged. The additional 2.7s per generator update comprises approximately 1.6s for VAE re-encoding, 0.9s for the counterfactual forward, and 0.2s for motion-proxy computation and regularization backward.

\section{Additional Quantitative Results}
\label{sec:additional_quant}
\subsection{Generation at the Training Horizon}
\label{sec:5s_results}
\cref{tab:5s_results} compares Self-Forcing and its two variants at the 5s training horizon, without temporal extrapolation. Frame Sink reduces Dynamic Degree from 65.00 to 37.00. Disen-Forcing increases it to 74.00 while preserving comparable scores on the other five dimensions.

\begin{center}
\begin{minipage}{\linewidth}
\centering
\captionof{table}{\textbf{Six-dimensional VBench results at the 5s training horizon.} All scores are percentages; higher is better.}
\label{tab:5s_results}
\small
\setlength{\tabcolsep}{5pt}
\begin{tabular}{lrrrrrr}
\toprule
Method & Subject & Background & Smoothness & Dynamic & Aesthetic & Imaging \\
\midrule
Self-Forcing & 93.81 & 94.59 & 98.29 & 65.00 & 63.33 & 69.53 \\
+ Frame Sink & 96.17 & 95.72 & 98.94 & 37.00 & 64.19 & 69.90 \\
+ Disen-Forcing & 94.17 & 94.79 & 98.23 & 74.00 & 63.87 & 70.27 \\
\bottomrule
\end{tabular}
\end{minipage}
\end{center}

\subsection{VideoScore Evaluation}
\label{sec:videoscore}
\begin{figure}[ht]
    \centering
    \includegraphics[width=\linewidth]{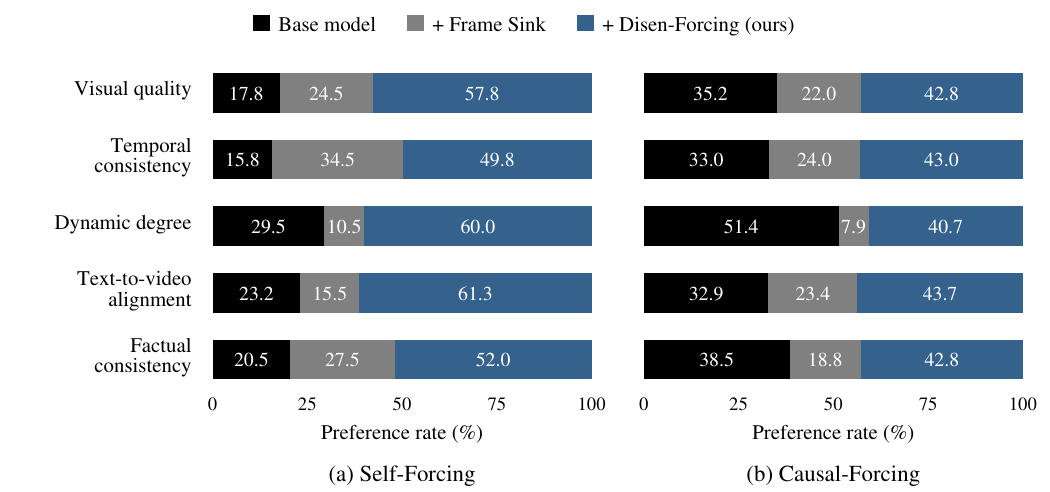}
    \caption{\textbf{VideoScore preference rates on 30s generations.} Percentages compare the three methods within each base model and evaluation dimension.}
    \label{fig:videoscore}
\end{figure}

We complement VBench-Long with VideoScore~\cite{he-etal-2024-videoscore}, a video evaluator trained on human quality ratings. For each base model, we compare the original model, its Frame Sink variant, and its Disen-Forcing variant on 30s generations for the 200 evaluation prompts. \cref{fig:videoscore} reports preference rates across five dimensions. Disen-Forcing has a higher preference rate than Frame Sink in every dimension on both base models. For Causal-Forcing, the base model has the highest dynamic-degree preference rate, while Disen-Forcing leads in the other four dimensions.

\section{Additional Qualitative Results}
\label{sec:qual}

\subsection{Anchor Head Masking Experiment}

\paragraph{Masking protocol.} To probe the functional role of \anchorheads{} during inference, we run the trained Self-Forcing checkpoint with three masking conditions: (1) \textit{no mask} (the unmodified rollout), (2) \textit{random mask} (zero out the keys/values of a randomly selected 20\% of heads in each layer, excluding identified anchor heads), and (3) \textit{anchor mask} (zero out the keys/values of the heads identified as anchors in $\cA$ (\cref{sec:contextsep})). To isolate the role of anchor heads in the long-range anti-drift mechanism rather than in within-chunk denoising, masking is applied only to the \emph{earliest-resident} KV frame in the sliding window, and only for AR rollout steps beyond the first chunk. All other settings, including the prompt and random seed, are identical across conditions.

\begin{figure}[h]
    \centering
    \includegraphics[width=\linewidth]{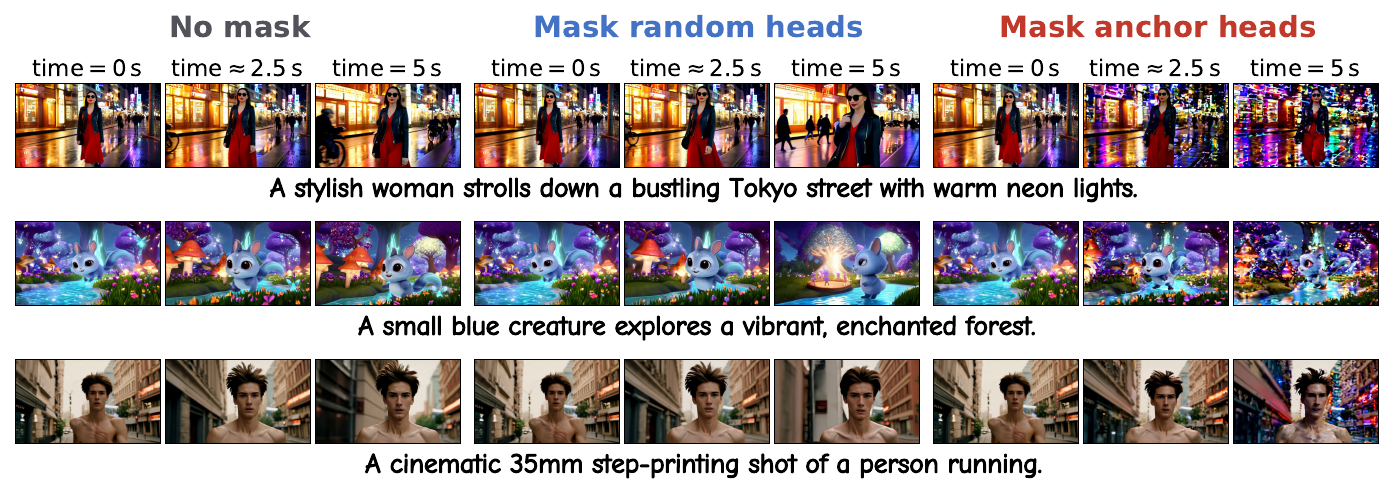}
    \caption{\textbf{Qualitative anchor-head masking comparison across prompts.}
    Each row shows one prompt rendered under three masking conditions: \textit{no mask} (baseline, left block), \textit{mask random heads} (control, middle block, \textcolor{workingblue}{\textbf{blue}}), and \textit{mask anchor heads} (treatment, right block, \textcolor{anchorred}{\textbf{red}}). Each block displays three frames spanning the 5-second clip. A brief prompt description is shown below each row. Random masking preserves visual fidelity; anchor masking produces visible identity, color, and scene drift, confirming that long-range anti-drift is carried specifically by the anchor-head subset.}
    \label{fig:head_masking_qual}
\end{figure}

As shown in \cref{fig:head_masking_qual}, randomly masking working heads produces videos with similar visual quality to the no-mask baseline, while anchor masking induces visible appearance drift---color shifts, identity inconsistency, and scene degradation---even within the 5-second training horizon. This pattern complements the quantitative VBench-Long results in \cref{fig:mask_head} and confirms that the anti-drift role is specifically carried by the sparse anchor-head subset rather than by attention heads in general.

\subsection{More Qualitative Long-Video Generation Results}
\label{sec:qual_long}

\begin{figure}[h]
    \centering
    \includegraphics[width=\linewidth]{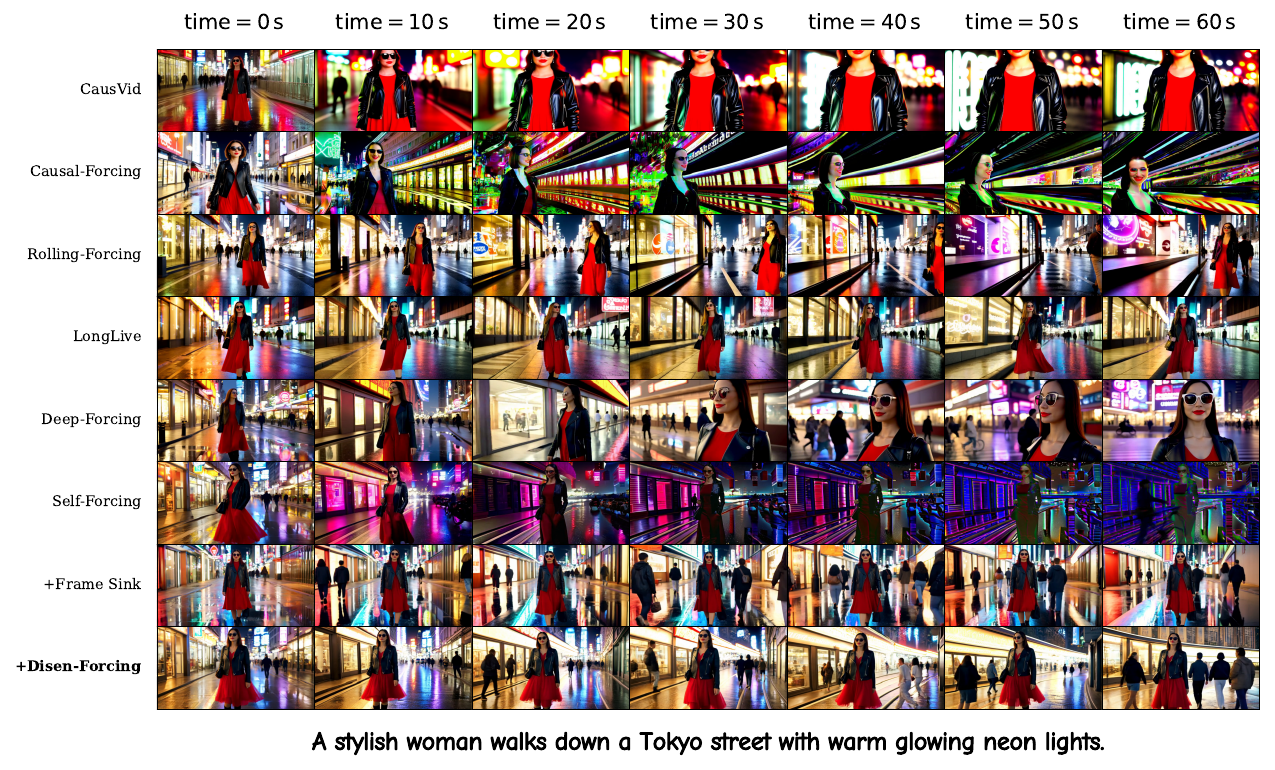}
    \caption{\textbf{1-minute qualitative comparison.}}
    \label{fig:supp_qual_1min_woman}
\end{figure}

\begin{figure}[h]
    \centering
    \includegraphics[width=\linewidth]{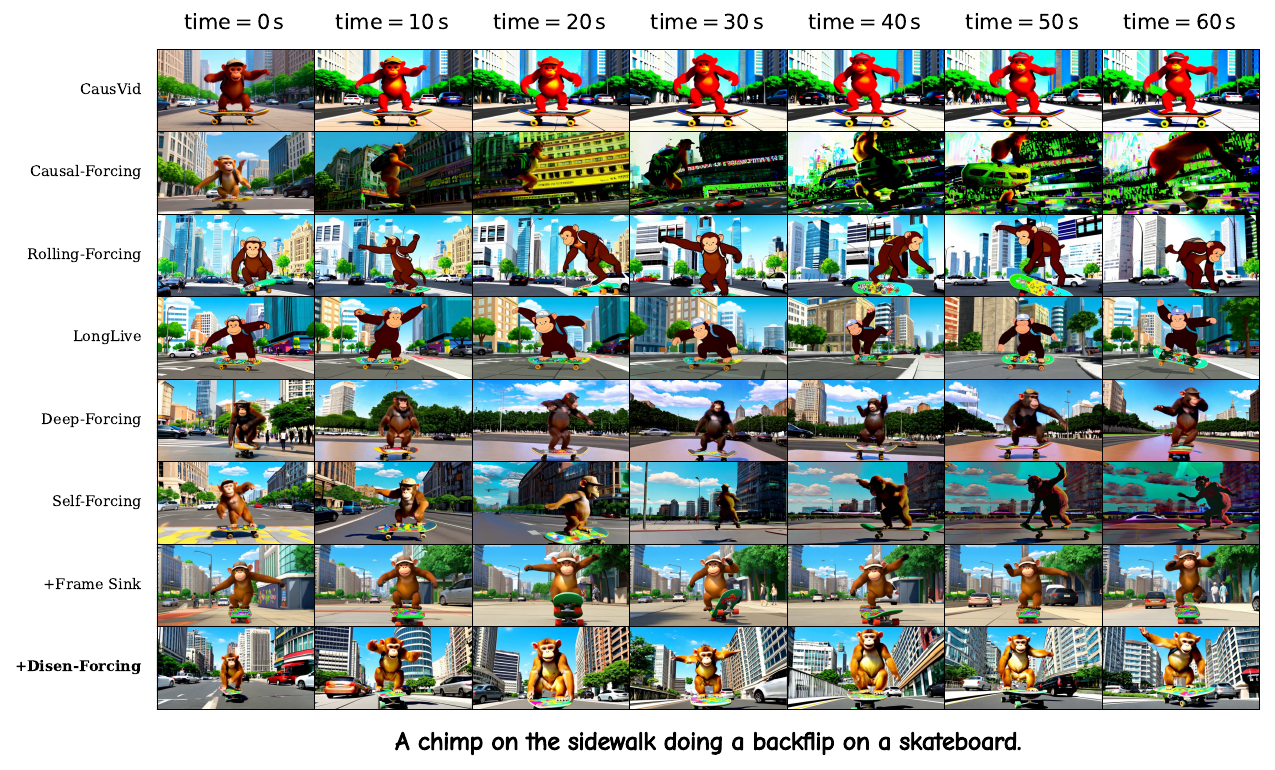}
    \caption{\textbf{1-minute qualitative comparison.}}
    \label{fig:supp_qual_1min_chimp}
\end{figure}

\begin{figure}[h]
    \centering
    \includegraphics[width=\linewidth]{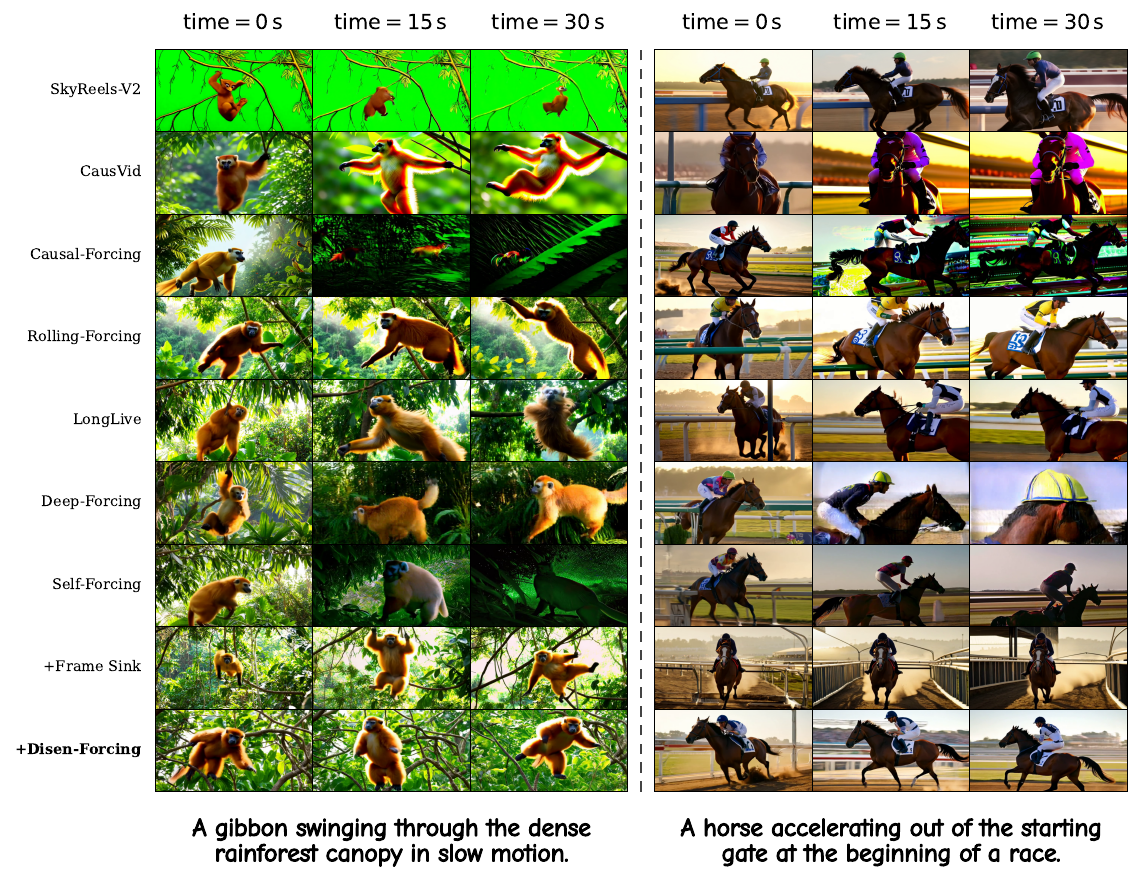}
    \caption{\textbf{30-second qualitative comparison.}}
    \label{fig:supp_qual_30s_gibbon_horse}
\end{figure}

\begin{figure}[h]
    \centering
    \includegraphics[width=\linewidth]{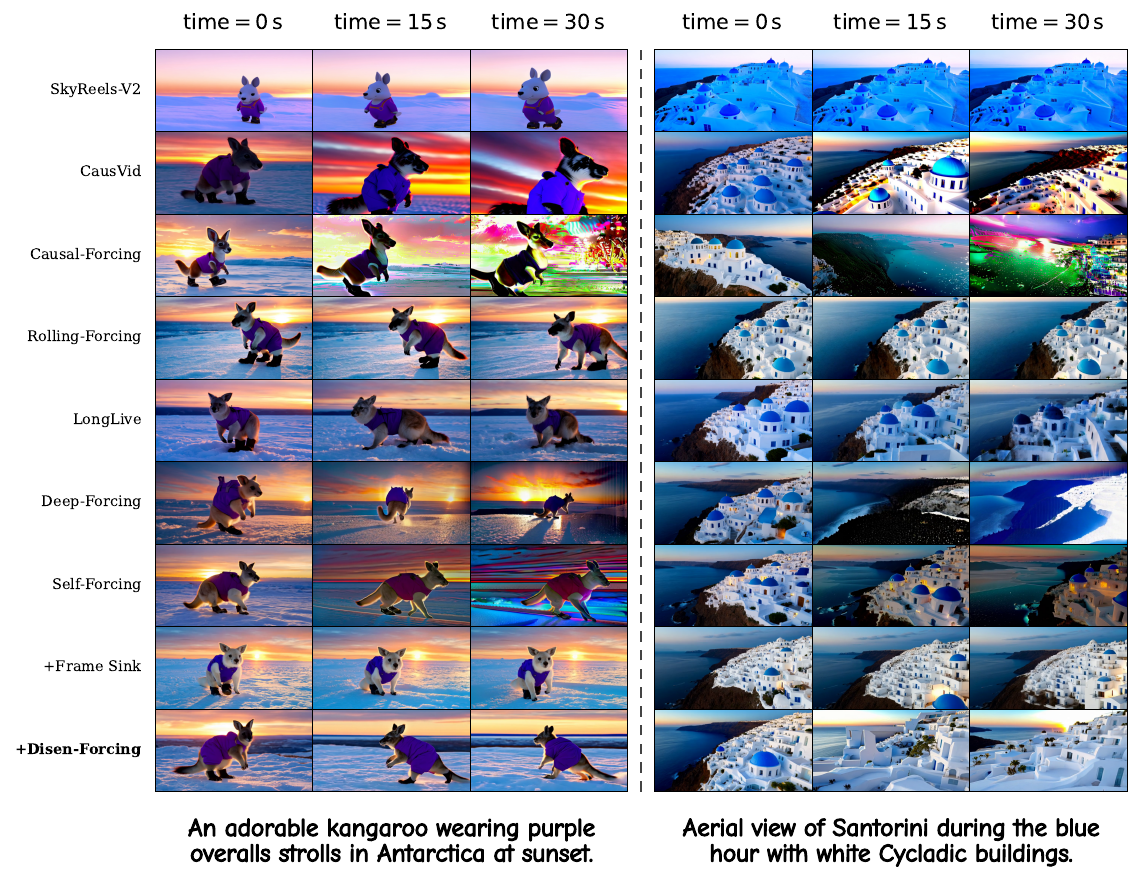}
    \caption{\textbf{30-second qualitative comparison.}}
    \label{fig:supp_qual_30s_kangaroo_santorini}
\end{figure}

We provide additional 30s and 1min qualitative comparisons across the baselines from main paper. \cref{fig:supp_qual_1min_woman,fig:supp_qual_1min_chimp} cover the 1-minute setting, and \cref{fig:supp_qual_30s_gibbon_horse,fig:supp_qual_30s_kangaroo_santorini} cover the 30-second setting. Across prompts, sliding-window methods drift visibly past the training horizon, sink-based methods preserve appearance but suppress motion, and \textbf{+Disen-Forcing} maintains both consistency and motion dynamics.

\section{Discussion: When Does Head-Type Separation Emerge?}
\label{sec:head_emergence}

\begin{figure}[h]
    \centering
    \includegraphics[width=\linewidth]{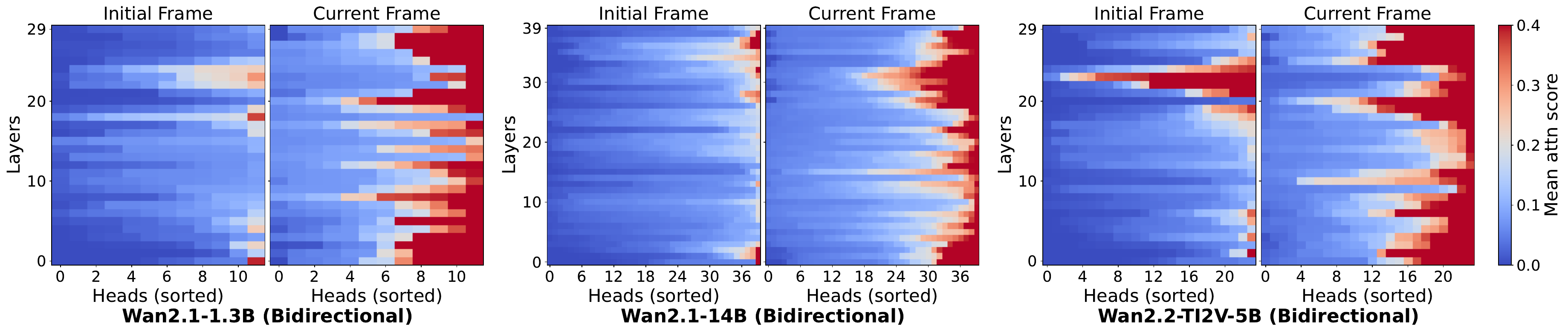}
    \caption{\textbf{Per-head temporal attention patterns in bidirectional Wan base models.}
    Attention scores averaged over 50 prompts and diffusion timesteps for the pretrained bidirectional
    Wan2.1-1.3B (left, 30 layers, 12 heads), Wan2.1-14B (middle, 40 layers, 40 heads), and
    Wan2.2-TI2V-5B (right, 30 layers, 24 heads), \emph{before} any causal AR training or distillation.
    All three models exhibit the same structural pattern observed in the AR-VDM variants of
    \cref{fig:sink_value}: most heads attend predominantly to the current frame, while a sparse subset
    at specific layers shows concentrated attention to the initial frame.}
    \label{fig:base_model_attn}
\end{figure}

We investigate \emph{when} the anchor/working head-type partition arises by measuring the
same temporal attention distribution in pretrained bidirectional Wan backbones, prior to any
causal AR adaptation or distillation.

\newparagraph{The partition is a structural property of the pretrained backbone.}
As shown in \cref{fig:base_model_attn}, the bidirectional Wan2.1-1.3B, Wan2.1-14B, and
Wan2.2-TI2V-5B base models~\cite{wan2025wan} all exhibit the same characteristic pattern: the vast majority
of heads attend predominantly to the current frame, while a sparse, stable subset shows
concentrated attention to the initial frame. This is consistent with the pattern observed
in the distilled AR-VDM variants of \cref{fig:sink_value}. The anchor/working partition
is therefore \emph{not induced by causal AR training or DMD distillation} --- it is
inherited from the pretrained backbone and further reinforced during distillation.

\newparagraph{The pattern scales to larger models.}
The same structural separation is observed in the 14B parameter variant of Wan2.1
(\cref{fig:base_model_attn}, middle): a sparse subset of heads shows elevated initial-frame
attention at both the earliest layers (0--2) and the deepest layers (28, 32, 36--38), while
the majority attend predominantly to the current frame. The pattern further transfers to
Wan2.2-TI2V-5B (\cref{fig:base_model_attn}, right) despite substantial architectural changes
(per-token time modulation, unified cross-attention, $4{\times}16{\times}16$ high-compression
VAE). This suggests that the head-type partition is a robust property across both model
scales \emph{and} Wan generations, and that Disen-Forcing's disentangled routing and
regularization can in principle be applied to more powerful AR-VDMs in future work.

\newparagraph{The partition extends beyond Wan.}
We additionally analyze HunyuanVideo-13B~\cite{hunyuan} and CogVideoX-5B~\cite{yang2024cogvideox}. \cref{tab:backbone_partition} summarizes two signatures of the partition: the fraction of heads attending more to the current frame than to the initial frames, and the ratio of mean initial-frame attention mass between anchor and working heads. We identify anchor heads using the same initial-frame attention criterion. Both models exhibit a majority of locally attending heads and stronger initial-frame attention in the anchor subset (2.4--3.6$\times$), consistent with the Wan backbones.

\begin{center}
\begin{minipage}{\linewidth}
\centering
\captionof{table}{\textbf{Head partition across pretrained backbones.} Local-head fraction is the percentage of heads with greater attention to the current frame than to the initial frames. Attention ratio is the mean initial-frame attention mass of anchor heads divided by that of working heads.}
\label{tab:backbone_partition}
\small
\begin{tabular}{lrr}
\toprule
Backbone & Local-head fraction (\%) & Attention ratio \\
\midrule
Wan2.1-1.3B & 76.7 & 4.2$\times$ \\
Wan2.1-14B & 69.2 & 3.8$\times$ \\
Wan2.2-TI2V-5B & 73.6 & 7.4$\times$ \\
HunyuanVideo-13B & 83.3 & 3.6$\times$ \\
CogVideoX-5B & 77.3 & 2.4$\times$ \\
\bottomrule
\end{tabular}
\end{minipage}
\end{center}

\newparagraph{AR distillation strengthens anchor-head attention to initial frames.}
Comparing the ODE-initialization and post-DMD panels in \cref{fig:sink_value}
reveals a consistent trend: DMD distillation increases the attention score of anchor heads to
the initial frames. For Self-Forcing, the mean anchor-head initial-frame attention rises from
$0.121$ (ODE init) to $0.150$ after DMD (+24\%). For Causal-Forcing, which starts from a
stronger ODE initialization, the anchor attention is already higher ($0.185$) and the gain
from distillation is correspondingly smaller ($0.194$, +5\%). This suggests that AR training
progressively sharpens the initial-frame anchoring role of the sparse anchor-head subset to support better streaming generation.

\section{Ablation Studies}
\label{sec:full_ablation}

\begin{table}[!ht]
\caption{\textbf{Full ablation results.} All six VBench-Long dimensions, their mean (Total), and quality drift for the variants in \cref{tab:ablation}, the regularization start step, regularizer implementation, and sink length.}
\label{tab:ablation_full}
\centering\small
\setlength{\tabcolsep}{3pt}
\renewcommand{\arraystretch}{1.08}
\begin{adjustbox}{max width=\linewidth}
\begin{tabular}{@{}l cccccccc@{}}
\toprule
& \multicolumn{7}{c}{VBench-Long Scores\,$\uparrow$} & \multirow{2}{*}{$\Delta^{\mathrm{Q}}_{\mathrm{Drift}}$\,$\downarrow$} \\
\cmidrule(lr){2-8}
Setting & Total & Subject & Background & Smooth & Dynamic & Aesthetic & Imaging & \\
\midrule
\rowcolor{groupgray}
\multicolumn{9}{@{}l}{\textit{Context routing and regularization}} \\
Self-Forcing + Frame Sink & 76.20 & 97.25 & 95.81 & 98.33 & 34.69 & 60.34 & 70.76 & 4.40 \\
\quad + Routing & 79.17 & 96.93 & 95.97 & 98.48 & 52.56 & 61.24 & 69.82 & 4.08 \\
\quad + Routing \& $\mathcal{L}_{\mathrm{reg}}$ & 82.45 & 95.96 & 95.17 & 97.36 & 77.19 & 60.06 & 68.97 & 4.39 \\
\midrule
\rowcolor{groupgray}
\multicolumn{9}{@{}l}{\textit{Impact of $\tau_p$}} \\
$p{=}70$ & 78.99 & 97.70 & 96.32 & 98.59 & 48.22 & 62.22 & 70.90 & 3.93 \\
$p{=}80$ & 82.45 & 95.96 & 95.17 & 97.36 & 77.19 & 60.06 & 68.97 & 4.39 \\
$p{=}90$ & 83.45 & 95.64 & 94.91 & 97.28 & 83.47 & 60.86 & 68.57 & 4.77 \\
\midrule
\rowcolor{groupgray}
\multicolumn{9}{@{}l}{\textit{Anchor-head set $\cA$ selection}} \\
Identified anchor heads & 82.38 & 96.52 & 95.50 & 98.13 & 73.12 & 60.95 & 70.05 & 4.03 \\
Layer-matched random heads & 79.20 & 97.45 & 96.09 & 98.47 & 51.16 & 61.94 & 70.08 & 4.51 \\
Globally random heads & 76.96 & 98.15 & 97.04 & 98.88 & 34.69 & 62.35 & 70.67 & 3.25 \\
\midrule
\rowcolor{groupgray}
\multicolumn{9}{@{}l}{\textit{Regularization layer $\mathcal{S}$ selection}} \\
Layer~20 & 82.35 & 96.20 & 95.22 & 97.85 & 75.66 & 59.57 & 69.59 & 5.01 \\
Layer~26 & 82.38 & 96.52 & 95.50 & 98.13 & 73.12 & 60.95 & 70.05 & 4.03 \\
Layer~27 & 81.18 & 96.68 & 95.76 & 97.85 & 65.56 & 61.55 & 69.69 & 4.26 \\
Layer~28 & 80.88 & 96.48 & 95.56 & 97.95 & 64.81 & 61.40 & 69.06 & 4.18 \\
Layers 20, 26, 27, 28 & 82.72 & 95.78 & 95.10 & 97.43 & 78.38 & 60.68 & 68.92 & 4.83 \\
\midrule
\rowcolor{groupgray}
\multicolumn{9}{@{}l}{\textit{Impact of $\lambda_{\mathrm{reg}}$}} \\
$\lambda_{\mathrm{reg}} = 0.001$ & 80.22 & 97.07 & 95.94 & 98.31 & 58.47 & 61.33 & 70.21 & 4.72 \\
$\lambda_{\mathrm{reg}} = 0.002$ & 82.38 & 96.52 & 95.50 & 98.13 & 73.12 & 60.95 & 70.05 & 4.03 \\
$\lambda_{\mathrm{reg}} = 0.005$ & 82.53 & 96.24 & 95.46 & 97.92 & 75.97 & 60.69 & 68.90 & 4.47 \\
\midrule
\rowcolor{groupgray}
\multicolumn{9}{@{}l}{\textit{Regularization start step ($\lambda_{\mathrm{reg}}{=}0.005$)}} \\
0 & 83.48 & 95.41 & 94.82 & 97.22 & 85.25 & 59.43 & 68.78 & 4.93 \\
50 & 80.39 & 96.58 & 95.55 & 98.04 & 61.91 & 60.60 & 69.67 & 4.51 \\
100 & 82.53 & 96.24 & 95.46 & 97.92 & 75.97 & 60.69 & 68.90 & 4.47 \\
150 & 81.94 & 96.29 & 95.33 & 97.93 & 72.66 & 60.60 & 68.83 & 4.86 \\
200 & 80.52 & 96.89 & 95.83 & 98.36 & 61.06 & 61.34 & 69.64 & 4.32 \\
\midrule
\rowcolor{groupgray}
\multicolumn{9}{@{}l}{\textit{Regularizer implementation}} \\
Full configuration & 82.38 & 96.52 & 95.50 & 98.13 & 73.12 & 60.95 & 70.05 & 4.03 \\
w/o adjacent matching & 83.33 & 96.14 & 95.27 & 97.64 & 81.16 & 60.64 & 69.11 & 4.19 \\
w/o VAE re-encoding & 82.35 & 96.09 & 95.21 & 97.89 & 75.53 & 60.02 & 69.37 & 4.85 \\
\midrule
\rowcolor{groupgray}
\multicolumn{9}{@{}l}{\textit{Sink length}} \\
1 & 82.74 & 95.09 & 94.54 & 97.34 & 82.25 & 58.82 & 68.40 & 5.09 \\
3 & 82.38 & 96.52 & 95.50 & 98.13 & 73.12 & 60.95 & 70.05 & 4.03 \\
6 & 80.75 & 96.80 & 95.81 & 98.12 & 63.81 & 60.82 & 69.13 & 4.59 \\
\bottomrule
\end{tabular}
\end{adjustbox}
\end{table}

Removing adjacent matching increases Dynamic Degree, while all five other dimensions decrease; removing VAE re-encoding increases $\Delta^{\mathrm{Q}}_{\mathrm{Drift}}$ from $4.03$ to $4.85$.

\subsection{Selecting the motion-proxy layer}\label{sec:proxy_layer}

Our motion-disentangling regularization
(\cref{sec:reg}) uses cross-frame attention maps from \workingheads{}-only layers. We consider layers 20, 26, 27, and 28, which contain no identified anchor heads.
We evaluate each candidate layer with a label-free reconstruction criterion:
we chain its cross-frame attention argmax across frames, warp
the reference frame along the resulting correspondence, and measure $L_1$
error against the decoded next frame. Normalizing by a no-motion
(identity-warp) baseline yields a relative score: below~$1$ means the
correspondence reconstructs inter-frame motion better than copying the frame.
Scores are averaged over 7 motion prompts~$\times$~2 seeds.

\begin{figure}[ht]
  \centering
  \includegraphics[width=\linewidth]{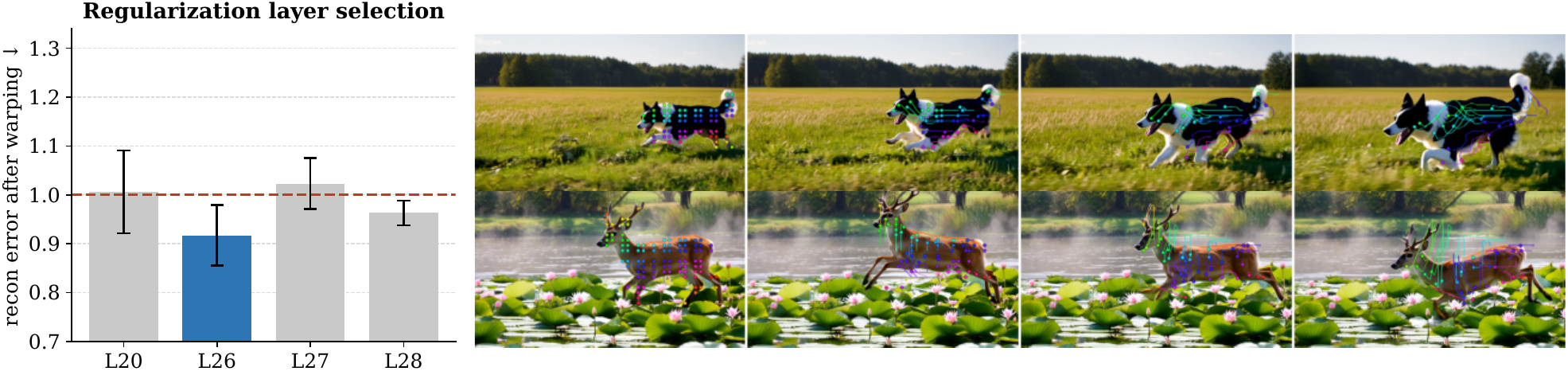}
  \caption{\textbf{Reconstruction-based motion-proxy layer analysis.}
    \emph{Left:} reconstruction error for candidate layers 20, 26, 27, and 28 from warping the reference
    frame along cross-frame attention chains, normalized to a no-motion
    baseline (mean$\pm$std over motion prompts$\times$seeds); layer~26 has
    the lowest error.
    \emph{Right:} example attention-based point tracks on a
    running dog (top) and a leaping deer (bottom).}
  \label{fig:proxy-selection}
\end{figure}

Among the four candidate layers, \textbf{layer~26} attains the lowest
reconstruction error (\cref{fig:proxy-selection}, left). We use layer~26 for motion-disentangling regularization. Qualitative point tracks
obtained by chaining cross-frame attention argmax through the rollout
illustrate subject motion and camera-induced background motion
(\cref{fig:proxy-selection}, right). The quantitative regularization-layer
ablation is reported in \cref{tab:ablation}.

\section{Discussions}
\paragraph{Limitations and Future Work}
\label{sec:limitations}
Disen-Forcing operates as a distillation-time plug-in and inherits the pretraining biases of the base model; it does not address artifacts originating from the bidirectional teacher itself. Its performance is further limited by the ODE initialization. The motion-disentangling regularization also requires a counterfactual forward pass during training, introducing a small per-step cost on the generator side (\cref{tab:efficiency}).
Several directions remain open. First, extending Disen-Forcing to the Wan2.1-14B and Wan2.2-TI2V-5B backbones, where the same anchor/working partition emerges (\cref{fig:base_model_attn}), and examining whether the routing principle generalizes to image-conditioned long-horizon generation. Second, while our instantiation builds on DMD distillation, extending the framework to other distillation objectives such as SiD-style losses~\cite{sid} or GAN-based objectives~\cite{goodfellow2020generative} is a natural next step. Finally, it would be worthwhile to investigate whether disentangling semantic anchoring from motion can also benefit the ODE initialization stage itself.

\paragraph{Broader Impacts}
Disen-Forcing improves the visual quality and long-range consistency of long-horizon video generation, which has applications in education, accessibility, and creative tools, but shares standard dual-use considerations of generative video models (e.g., misuse for synthetic media). The method is a fine-tuning plug-in on existing public AR-VDMs and does not in itself introduce new generative capability beyond those baselines.

\end{document}